\documentclass{article}

\usepackage[super,sort,comma,numbers]{natbib}
\usepackage[accepted]{icml2026}
\AtBeginDocument{%
  \bibpunct{}{}{,}{s}{}{}%
  \citestyle{nature}%
}

\usepackage{microtype}
\usepackage{graphicx}
\usepackage{subcaption}
\usepackage[table]{xcolor}
\usepackage{booktabs} % for professional tables
\usepackage{multirow} % for multirow cells in tables
\usepackage{tabularx}
\usepackage{comment}

\usepackage{url}

\usepackage{amsmath}
\usepackage{amssymb}
\usepackage{mathtools}
\usepackage{algorithm}

\usepackage{hyperref}
\makeatletter
\AtBeginDocument{%
  \@ifundefined{theHalgorithm}{%
  }{%
  }%
}
\makeatother

\makeatletter
\renewcommand{\ICML@appearing}{} 
\renewcommand{\ICML@preprint}{}  
\makeatother

\begin{document}

\hbadness=10000
\vbadness=10000

\twocolumn[
\icmltitle{Revealing Training Data Exposure in Vision–Language Large Models via Parameter Gradients}

\begin{icmlauthorlist}
\icmlauthor{Zhihao Zhu}{aff1}
\icmlauthor{Hongyi Tang}{aff1}
\icmlauthor{Yi Yang}{aff1}
\icmlauthor{Ahmed Abbasi}{aff2}
\end{icmlauthorlist}

\icmlaffiliation{aff1}{Department of Information Systems, Business Statistics and Operations Management (ISOM), Hong Kong University of Science and Technology, Hong Kong, China.}
\icmlaffiliation{aff2}{Department of IT, Analytics, and Operations, University of Notre Dame, Notre Dame, Indiana, USA} 
\icmlcorrespondingauthor{Yi Yang}{imyiyang@ust.hk}
\icmlkeywords{Vision-Language Large Models, Copyright verification, Training data detection}
\vskip 0.4in
]

\begin{abstract}
Vision-Language Large Models (VLLMs) trained on massive crawled corpora raise pressing copyright and data-provenance concerns. These concerns are particularly acute in healthcare, where patient medical images paired with clinical reports demand rigorous privacy safeguards. However, existing training data detection methods either fail in cross-modal scenarios or rely on superficial output signals with insufficient discriminative power. We introduce \textbf{GradAudit}, a gradient-based auditing framework that examines internal optimization dynamics rather than treating VLLMs as black boxes. Our approach builds on a key observation: model parameters converge to regions where gradients on training samples become stable and well-aligned, whereas gradients on non-training samples remain noisy and inconsistent. By analyzing these gradient signatures, GradAudit achieves strong separability and detects genuine image-text associations learned during training, not merely individual modality membership. Empirically, across both medical and general-domain datasets, GradAudit substantially outperforms state-of-the-art baselines in both pretraining and fine-tuning VLLMs. In a case study employing copyrighted content, we show that existing training data detection methods not only underestimate the extent of unauthorized data usage, but that this underestimation becomes more pronounced as models become more recent and more advanced.
\end{abstract}

\printAffiliationsAndNotice{}

\section*{Introduction}
\label{sec:introduction}
The rapid advancement of artificial intelligence has catalyzed transformative developments across diverse applications~\citealp{wang2023scientific, AI_research_impact_2025}. Vision-Language Large Models (VLLMs)~\citealp{zhuminigpt, zhang2024vision}, which integrate visual and textual modalities to achieve cross-modal understanding and generation, have emerged as a cornerstone technology driving this transformation. By learning from vast multimodal datasets, VLLMs demonstrate remarkable capabilities in tasks ranging from image captioning~\citealp{hu2022scaling} and visual question answering~\citealp{chappuis2022prompt} to multimodal content generation~\citealp{wu2024visionllm}.

However, the widespread deployment of VLLMs has raised critical concerns regarding data copyright and provenance~\citealp{dataguard2024privacy}. The multimodal nature of these systems amplifies infringement risks, as training requirements span images, text, and potentially other data types simultaneously. These concerns are particularly serious in healthcare, where patient data are protected by stringent legislative frameworks such as the Health Insurance Portability and Accountability Act (HIPAA), and violations involving unauthorized use can result in severe legal consequences. For example, a class-action lawsuit was filed against the University of Chicago Medical Center for providing thousands of patients' un-anonymized electronic health records for Google's AI system development~\citealp{wakabayashi2019google}. Beyond the healthcare domain, similar concerns have emerged in the general AI industry. For example, the use of “Ghibli-style” image generation has drawn public criticism from creators~\citep{ghibli2025lawsuit}, while Getty Images has filed a lawsuit against Stability AI, alleging that millions of its copyrighted images were used without authorization for model training~\citealp{getty2025stability}. These cases highlight an urgent need for auditing tools that can detect unauthorized usage of multimodal data in AI models ~\citealp{zhou2023unified, felzmann2020towards}.

A fundamental question emerges: \textit{how can we determine whether a deployed VLLM has been trained on specific  vision-language data?} For example, given medical images and their associated diagnostic records, how can we verify whether a VLLM has used these image-text pairs during training? This problem, known as training data detection (or equivalently, data auditing), represents an essential challenge for responsible AI deployment and compliance verification.

However, current approaches exhibit significant limitations when applied to VLLMs. First, most existing methods target single modalities (analyzing token-level probabilities for text models~\citealp{shi2024detecting} or output variations for vision models~\citealp{wen2023canary}). These modality-specific designs are fundamentally unsuitable for cross-modal VLLMs. Second, the few methods proposed for VLLMs rely on superficial signals such as output entropy~\citealp{li2024membership} without examining internal model dynamics. As we demonstrate empirically, these approaches achieve near-random performance (50--55\% AUROC) on most configurations, with insufficient discriminative power for copyright verification contexts.

To address these limitations, we propose \textbf{GradAudit}, the first gradient-based data auditing framework specifically designed for VLLMs. In contrast to methods that rely on token-level loss or augmented images, our approach leverages gradients, a signal shared across all components in VLLMs (e.g., vision encoders, language models, and cross-modal fusion modules), enabling effective utilization of information from different modalities. Our approach is grounded in a key observation: training data leave distinctive gradient signatures in the optimized model. During optimization, model parameters converge to a region where gradients on training samples become stable and well-aligned, whereas gradients on non-training samples remain noisy and inconsistent~\citealp{feng2021inverse, nasr2018comprehensive}. This asymmetry provides a powerful discriminative signal for detecting training data membership. GradAudit operates under a white-box setting, requiring access to model parameters. This setting is particularly relevant for internal audits and regulatory inspections, as regulations such as the Digital Services Act (DSA) explicitly require model providers to grant the necessary access for compliance assessment.

To handle the computational challenges posed by billions of parameters in modern VLLMs, GradAudit decomposes parameter gradient matrices into functionally interpretable slices along row and column dimensions, transforming gradient responses into feature vectors that capture specific functional components such as spatial encoding or semantic reasoning. However, not all functional slices exhibit uniform sensitivity to training data: generic components demonstrate low gradient sensitivity while specialized components show high sensitivity. To address this heterogeneity, we introduce a noise masking mechanism that leverages reference data comprising both training and non-training samples to systematically identify and suppress insensitive features. The final decision compares the masked gradient similarity between audited data and the reference training baseline, where higher similarity provides stronger evidence of training data membership.

Empirical evaluation across seven experimental configurations demonstrates that GradAudit substantially outperforms all baselines. Evaluated on vision-language benchmarks spanning both medical (PMC-OA, ROCO, MedTrinity) and general domains (COCO, FashionGen), GradAudit achieves AUROC of 87.2\% on pre-training scenarios and up to 92.7\% on fine-tuning scenarios, substantially outperforming all baselines. Crucially, GradAudit captures genuine image-text associations rather than individual modality membership: it achieves 85.5\% AUROC in distinguishing correctly paired training data from shuffled pairs where both images and texts appeared during training but not as pairs, a capability essential for multimodal copyright verification. Beyond controlled benchmarks, we conduct a case study examining copyrighted Studio Ghibli content in deployed VLLMs. 
GradAudit estimates roughly three times the exposure ratio detected by baseline methods, highlighting a substantial underestimation of unauthorized data usage in more recent and more advanced VLLMs.

Our work carries significant implications for governance and ethical deployment of AI systems. For regulators, GradAudit provides a detection tool that enables data compliance auditing. For content creators and rights holders, our method enables verification of whether copyrighted multimodal content has been incorporated into AI systems without authorization, thereby strengthening intellectual property protection in the AI era. For AI developers, GradAudit offers a mechanism to demonstrate compliance with data usage agreements, potentially mitigating legal risks and building stakeholder trust. By establishing reliable data auditing mechanisms, our work helps build trust among content creators, AI developers, and regulators, ultimately promoting transparency and accountability in AI development and deployment.

\section*{Results}
\label{sec:experiments}
\begin{table*}[htbp]
\centering
\caption{\textbf{Overview of seven experiment configurations.}
For models with explicitly documented training and test splits (e.g., BLIP-ITM on COCO), we use the training split as \textbf{Training Data} and the test split as \textbf{Non-training Data}. For models with only documented training datasets (e.g., BiomedCLIP on PMC-OA), we use distributionally similar datasets as \textbf{Non-training Data} (e.g., ROCO, which shares the same radiology image-text domain as PMC-OA).}
\label{tab:config}
\resizebox{0.85\textwidth}{!}{%
\begin{tabular}{@{}cccccccc@{}}
\toprule
\textbf{Exp.} & \textbf{Model}  & \textbf{Architecture} & \textbf{Params} & \textbf{Training Stage} & \textbf{Training Data} & \textbf{Non-training Data} \\ 
\midrule
1 & BiomedCLIP & Contrastive & 200M & Pretrain    & PMC-OA     & ROCO \\
2 & PubmedCLIP & Contrastive & 151M & Pretrain    & ROCO     & ROCO \\
3 & BLIP-ITM   & Retrieval   & 200M & Pretrain    & COCO       & COCO \\
4 & LLaVA-Med  & Generation  & 7.3B & Pretrain    & PMC-OA     & ROCO \\
5 & MiniGPT-v2  & Generation  & 8.2B & Pretrain    & COCO & COCO \\
6 & Qwen2-VL   & Generation  & 2.3B & Fine-tuning & MedTrinity   & MedTrinity \\
7 & Qwen2-VL   & Generation  & 2.3B & Fine-tuning & FashionGen   & FashionGen \\
\bottomrule
\end{tabular}%
}
\label{tab:models}
\end{table*}

\begin{table*}[t]
\centering
\caption{\textbf{Performance comparison across all experimental configurations.} We report AUROC (\%) and TPR@5\%FPR (\%) for each method. Bold values indicate the best performance per configuration. GradAudit consistently outperforms all baselines across diverse models and training scenarios.}
\label{tab:results-merged}
\footnotesize  
\setlength{\tabcolsep}{3pt} 
\begin{tabular}{l ccccccccccccccc}
\toprule
\multirow{3}{*}{\textbf{Method}} & 
\multicolumn{2}{c}{\textbf{BiomedCLIP}} &
\multicolumn{2}{c}{\textbf{PubmedCLIP}} &
\multicolumn{2}{c}{\textbf{BLIP-ITM}} & 
\multicolumn{2}{c}{\textbf{LLaVA-Med}} & 
\multicolumn{2}{c}{\textbf{MiniGPT-v2}} & 
\multicolumn{2}{c}{\textbf{Qwen-MedTrinity}} & 
\multicolumn{2}{c}{\textbf{Qwen-FashionGen}} \\
\cmidrule(lr){2-3} \cmidrule(lr){4-5} \cmidrule(lr){6-7} \cmidrule(lr){8-9} \cmidrule(lr){10-11} \cmidrule(lr){12-13} \cmidrule(lr){14-15}
& \multicolumn{1}{c}{\textbf{AUC}} & \multicolumn{1}{c}{\textbf{TPR}} & 
\multicolumn{1}{c}{\textbf{AUC}} & \multicolumn{1}{c}{\textbf{TPR}} &
\multicolumn{1}{c}{\textbf{AUC}} & \multicolumn{1}{c}{\textbf{TPR}} & \multicolumn{1}{c}{\textbf{AUC}} & \multicolumn{1}{c}{\textbf{TPR}} & \multicolumn{1}{c}{\textbf{AUC}} & \multicolumn{1}{c}{\textbf{TPR}} & \multicolumn{1}{c}{\textbf{AUC}} & \multicolumn{1}{c}{\textbf{TPR}} & \multicolumn{1}{c}{\textbf{AUC}} & \multicolumn{1}{c}{\textbf{TPR}} \\
\midrule
Loss        & 53.2 & 20.0 & 59.5 & 8.80 & 51.3 & 5.41  & 57.3 & 12.5  & 50.6 & 3.30 & 52.8 & 5.40  & 90.9 & 65.1 \\
Entropy     & 57.9 & 19.3 & 58.3 & 6.60 & 55.3 & 6.90  & 58.2 & 8.31  & 48.6 & 4.31 & 58.8 & 10.7  & 52.4 & 5.50 \\
Zlib        & 60.9 & 2.30 & 61.6 & 8.80 & 52.0 & 3.10  & 61.6 & 6.45  & 50.7 & 2.81 & 54.3 & 5.40  & 90.0 & 65.1 \\
Min-K       & 52.7 & 4.30 & 59.2 & 6.70 & 55.6 & 6.70  & 59.0 & 5.61  & 50.4 & 4.40 & 55.0 & 6.40  & 90.4 & 53.7 \\
Min-K++     & 53.5 & 23.5 & 60.7 & 10.9 & 54.3 & 6.38  & 59.3 & 1.40  & 51.1 & 6.12 & 50.9 & 4.20  & 87.7 & 42.4 \\
DC-PDD      & 53.2 & 11.9 & 61.6 & 8.50 & 54.3 & 1.90  & 63.9 & 12.9  & 50.1 & 5.10 & 55.8 & 6.40  & 57.7 & 8.71 \\
ModRényi    & 55.9 & 2.90 & 52.2 & 4.58 & 51.2 & 6.00  & 51.1 & 5.46  & 50.8 & 5.20 & 54.1 & 6.70  & 50.0 & 3.50 \\
Similarity  & 59.3 & 4.20 & 50.6 & 4.90 & 50.7 & 5.20  & 51.8 & 7.41  & 48.3 & 5.67 & 51.8 & 7.41  & 50.0 & 5.30 \\
M$^4$I      & 52.8 & 8.90 & 50.5 & 6.75 & 55.5 & 6.00  & 60.2 & 12.8  & 51.3 & 6.00 & 51.5 & 3.60  & 51.0 & 6.00 \\
GradNorm    & 50.5 & 17.3 & 52.7 & 5.75 & 59.3 & 10.9  & 43.2 & 3.10  & 50.4 & 5.50 & 77.8 & 11.2  & 78.8 & 49.2 \\
\midrule
\textbf{GradAudit} & \textbf{66.5} & \textbf{23.8} & \textbf{81.0} & \textbf{26.3} & \textbf{87.2} & \textbf{49.2} & \textbf{66.1} & \textbf{13.5} & \textbf{63.7} & \textbf{21.5} & \textbf{80.1} & \textbf{14.5} & \textbf{92.7} & \textbf{66.8} \\
\bottomrule
\end{tabular}
\label{tab:merged-results}
\end{table*}

\subsection*{Experimental Setup}
\noindent\textbf{VLLMs}. We evaluate GradAudit on six representative VLLMs spanning three architectural paradigms: BiomedCLIP~\citealp{zhang2023biomedclip} (contrastive), PubmedCLIP~\citealp{eslami2023pubmedclip} (contrastive), BLIP-ITM~\citealp{li2022blip} (retrieval), and LLaVA-Med~\citealp{li2023llava}, MiniGPT-v2~\citealp{zhuminigpt}, Qwen2-VL~\citealp{wang2024qwen2} (generative). These models cover diverse domains (medical and general), parameter scales (151M--8.2B), and training objectives, enabling comprehensive assessment of GradAudit's generalization capability.

\noindent\textbf{Benchmarks}. To construct auditing benchmarks, we collect datasets explicitly documented in the official training reports of evaluated VLLMs, including PMC-OA~\citealp{lin2023pmc} and  COCO~\citealp{lin2014microsoft}. For models with documented train/test splits (e.g., BLIP-ITM on COCO), we sample training data as positive examples and test data as negative examples. For models where only the training corpus is documented (e.g., BiomedCLIP on PMC-OA), we draw negative examples from distributionally similar datasets sharing the same domain characteristics (e.g., ROCO~\citealp{ruckert2024rocov2} for radiology image-text pairs). From each configuration, we sample 1,000 positive and 1,000 negative examples to construct balanced auditing benchmarks.

To further assess auditing performance under fine-tuning scenarios, the predominant paradigm in modern VLLM deployment, we conduct experiments on MedTrinity~\citealp{xiemedtrinity} and FashionGen~\citealp{rostamzadeh2018fashion}. Specifically, we randomly sample 1,000 examples from each dataset to fine-tune Qwen2-VL using LoRA~\citealp{hu2022lora}, treating these as positive samples. Another 1,000 randomly sampled examples, strictly excluded from fine-tuning, serve as negative samples. This configuration evaluates GradAudit's effectiveness in detecting fine-tuning data usage, precisely the scenario where copyright concerns are most acute.
Detailed experimental configurations are summarized in Table~\ref{tab:config}.

\noindent\textbf{Baselines}. We compare GradAudit against multiple baseline methods spanning different methodological approaches. Our baselines encompass two categories: (1) \textit{output-based methods} that examine model outputs, including Loss Attack~\citealp{yeom2018privacy}, Entropy~\citealp{song2021systematic}, Zlib~\citealp{carlini2021extracting}, Min-K~\citealp{shi2024detecting}, Min-K++~\citealp{zhang2025min}, DC-PDD~\citealp{zhang2024pretraining}, ModRényi~\citealp{li2024membership}, Similarity~\citealp{wu2025image}, and M$^4$I~\citealp{hu2022m}; and (2) \textit{gradient-based methods} such as GradNorm~\citealp{nasr2018comprehensive}, which directly aggregates gradient information. Notably, most baseline methods were originally designed for single-modality models. To our knowledge, M$^4$I, ModRényi and Similarity are the only existing data auditing methods specifically proposed for multimodal models. Detailed descriptions of these baseline methods are provided in the Appendix.

\noindent\textbf{Metrics}. We evaluate performance using two metrics commonly employed in prior training data detection work: AUROC (Area Under the Receiver Operating Characteristic curve) and TPR@5\%FPR (True Positive Rate at 5\% False Positive Rate). AUROC provides a comprehensive measure of overall classification performance across all possible thresholds. TPR@5\%FPR, in contrast, quantifies the detection rate under a fixed false positive rate of 5\%, which is particularly relevant for practical copyright verification scenarios where maintaining low false positive rates is essential to avoid erroneous accusations.

\subsection*{Overall Performance}
As shown in Table~\ref{tab:merged-results}, GradAudit consistently outperforms all baseline methods across all seven experimental configurations. In terms of AUC, GradAudit achieves scores ranging from 63.7\% to 92.7\%, substantially exceeding the best baseline in each case. For TPR@5\%FPR, GradAudit achieves values ranging from 13.5\% to 66.8\%, demonstrating strong detection capability under stringent false positive constraints. Notably, most baseline methods exhibit unstable performance across configurations, with many achieving AUROC scores close to 50.0\%, corresponding to random guessing. This instability is particularly pronounced for multimodal-specific methods such as M$^4$I, ModRényi and Similarity, which rely on assumptions that do not generalize well to diverse VLLM architectures and training objectives. For instance, on FashionGen, simple output-based methods (Loss, Zlib, Min-K) achieve approximately 90\% AUROC while all multimodal baselines perform at chance level. In contrast, GradAudit directly examines parameter gradients that reflect optimization dynamics regardless of architectural details, enabling consistent performance across diverse configurations.

Beyond the comparison with baselines, we observe substantial performance variations across different models, reflecting the heterogeneous nature of training configurations and their impact on data memorization patterns. Specifically, fine-tuning scenarios consistently demonstrate superior performance compared to pre-training settings. Fine-tuned Qwen2-VL achieves AUC scores of 80.1\% and 92.7\% on MedTrinity and FashionGen respectively, substantially higher than most pre-training configurations. This performance gap can be attributed to catastrophic forgetting in large-scale VLLM training: given the massive scale of pre-training data, early-stage training samples are more susceptible to forgetting during later training stages, making them less distinguishable from non-training data. In contrast, fine-tuning data, introduced after pre-training, exhibit stronger memorization and are thus more easily detected through data auditing. 

\begin{figure}[htbp]
\centering
\includegraphics[width=\columnwidth]{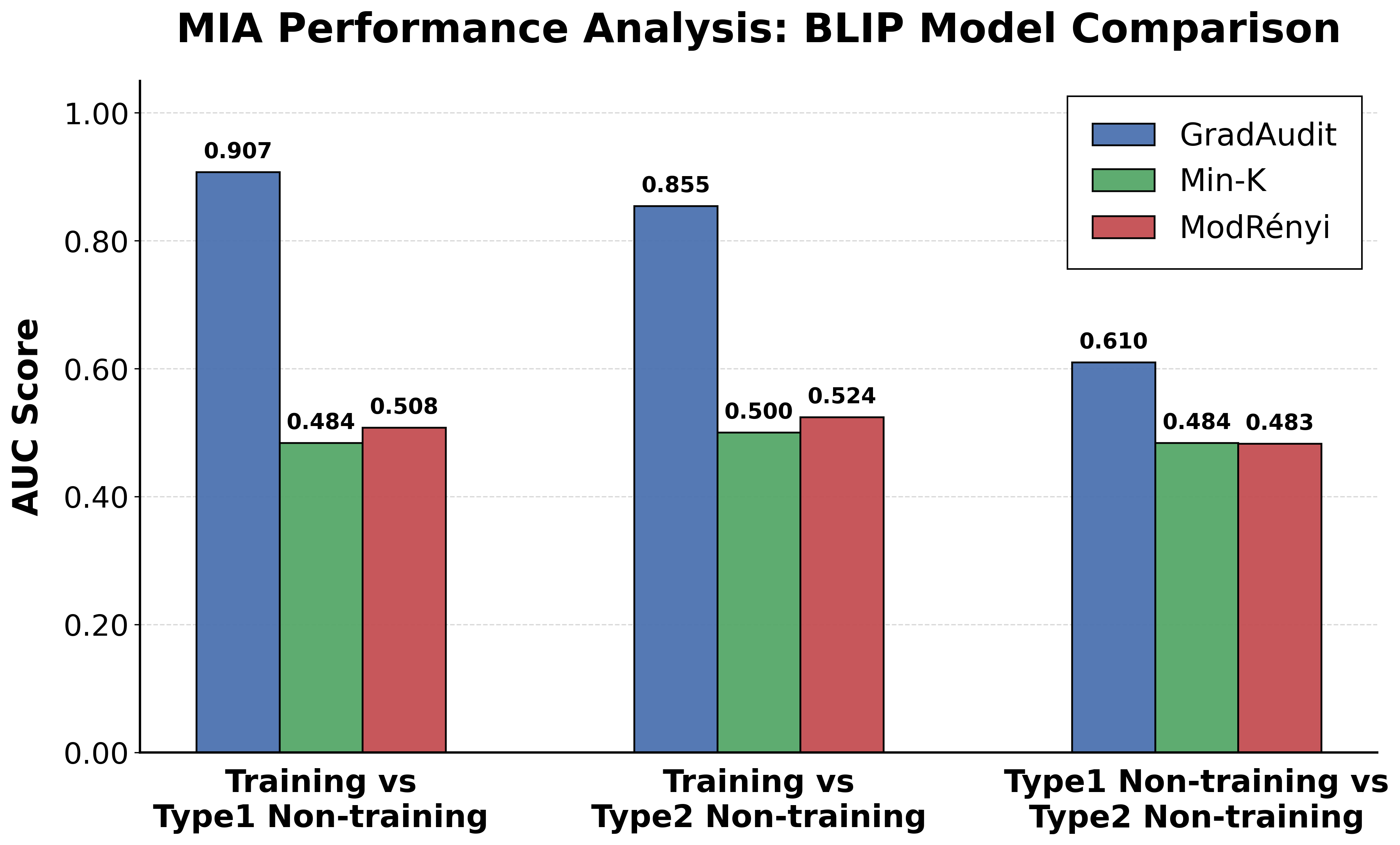}
\caption{\textbf{GradAudit detects image-text pairing relationships, not just individual data membership.} We evaluate three scenarios: (1) Training Data: correctly paired image-text from the training set; (2) Type-1 Non-training: held-out test data; (3) Type-2 Non-training: shuffled image-text pairs where both images and texts appeared in training, but not as pairs. GradAudit achieves 85.5\% AUROC in distinguishing Training Data from Type-2 Non-training, demonstrating its ability to capture joint multimodal representations rather than individual modality membership.}
\label{fig:pairing_analysis}
\end{figure}
\subsection*{In-depth Analysis}
\subsubsection*{Multimodal Pairing Auditing}
A critical question for multimodal data auditing is whether methods detect genuine image-text associations or merely individual modality membership. To investigate this, we design a controlled experiment with three data categories: (1) \textit{Training Data}, consisting of correctly paired image-text samples from the training set; (2) \textit{Type1 Non-training Data}, referring to held-out test data that the model has never encountered; and (3) \textit{Type2 Non-training Data}, comprising shuffled pairs created by randomly reassigning captions among training images. Importantly, both the images and texts in Type-2 individually appeared during training, but the specific pairings did not. Figure~\ref{fig:pairing_analysis} shows how GradAudit and state-of-the-art baselines distinguish between these data categories.

If a method merely detects whether individual images or texts appeared during training, it would fail to distinguish Training Data from Type-2 Non-training Data. However, GradAudit achieves 0.855 AUROC in this critical comparison, demonstrating that it captures joint multimodal representations rather than individual modality membership. Furthermore, GradAudit shows limited ability to distinguish Type-1 from Type-2 Non-training (0.610 AUROC), which is expected: both lack proper pairing relationships, so they produce similar gradient patterns. This result further confirms that GradAudit's discriminative signal stems from learned image-text associations.

\subsubsection*{Scaling Behavior of Data Auditing}
Scaling laws have emerged as fundamental principles governing language model behavior~\citealp{xiong2025temporal, kaplan2020scaling}. Here, we investigate whether analogous scaling relationships exist for data auditing by examining how auditing performance scales with both model size and fine-tuning data size. Understanding these relationships is crucial for predicting GradAudit's effectiveness on future, larger-scale VLLMs. To this end, we conduct controlled experiments on the Qwen2-VL model family using the MedTrinity dataset, systematically varying model size (2B, 3B, and 7B parameters) and fine-tuning data size (1K, 3K, and 5K samples).

\begin{figure*}[htbp]
\centering
\begin{subfigure}[b]{0.49\textwidth}
\centering
\includegraphics[width=\linewidth]{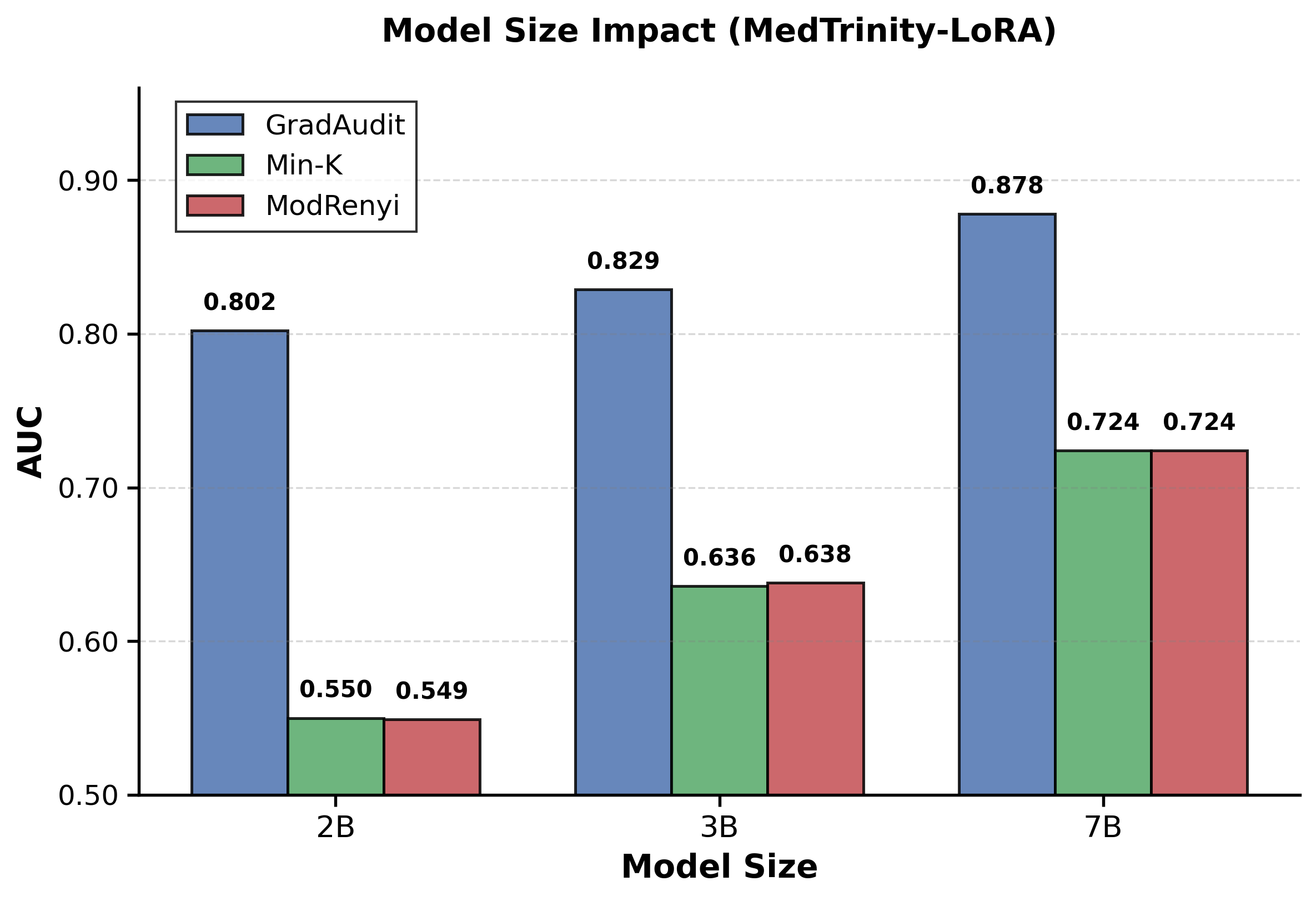}
\caption{\textbf{Scaling with model size.}}
\label{fig:lora_model_size}
\end{subfigure}\hfill
\begin{subfigure}[b]{0.49\textwidth}
\centering
\includegraphics[width=\linewidth]{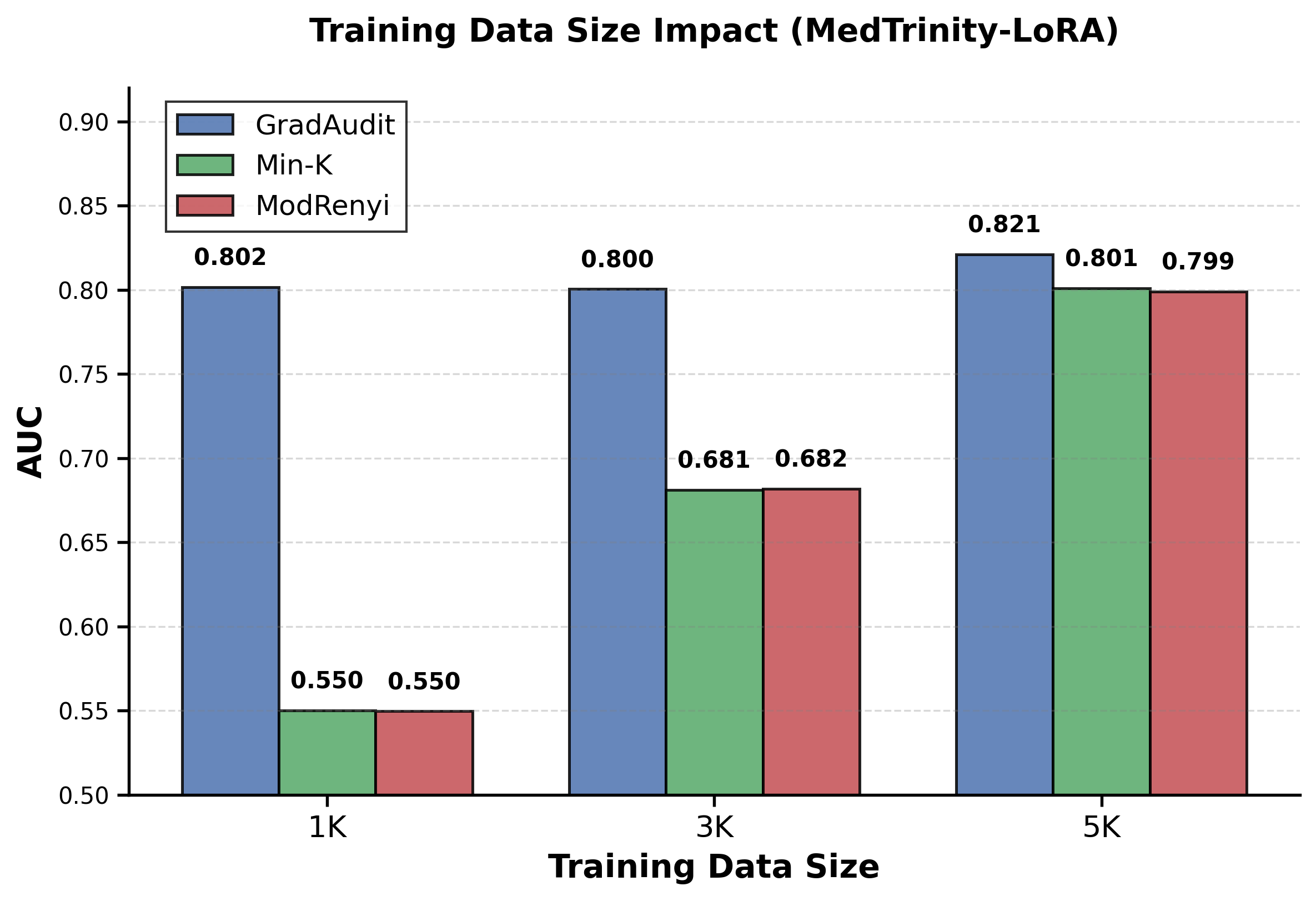}
\caption{\textbf{Scaling with training data size.}}
\label{fig:lora_training_size}
\end{subfigure}
\caption{\textbf{Scaling behavior of data auditing performance.} 
Auditing performance improves consistently with both model scale and fine-tuning data size, with GradAudit maintaining substantial advantages over baselines across all configurations.}
\label{fig:lora_scaling}
\end{figure*}

\textbf{Scaling with model size.}
As shown in Figure~\ref{fig:lora_model_size}, auditing performance improves consistently with model scale. GradAudit achieves AUROC of 0.802, 0.829, and 0.878 for 2B, 3B, and 7B models respectively, substantially outperforming the baseline ModRényi (0.550, 0.636, and 0.724). This finding aligns with prior research on training data detection~\citealp{shi2024detecting, tang2025identifying}: the fundamental mechanism underlying data auditing effectiveness is the model's memorization of training data, and this memorization capacity becomes more pronounced with increased model size. Notably, GradAudit maintains substantial advantages over baselines across all model sizes, with performance gaps ranging from 15.4 to 25.2 percentage points. This makes GradAudit particularly well-suited for modern AI systems where large-scale models have become mainstream, and suggests that GradAudit will remain effective as VLLMs continue to scale.

\textbf{Scaling with training data size.}
As shown in Figure~\ref{fig:lora_training_size}, auditing performance also improves with fine-tuning data size. GradAudit achieves AUROC of 0.802, 0.800, and 0.821 for 1K, 3K, and 5K fine-tuning samples respectively, maintaining a consistent advantage over baselines across all scales. This trend can be attributed to the domain shift between general-purpose pre-training and specialized fine-tuning data. Our fine-tuning datasets (e.g., MedTrinity for medical imaging) originate from specialized domains where samples share inherent correlations, while Qwen2-VL is pre-trained on general-purpose data. This domain shift triggers significant model adaptations, and as the fine-tuning data size increases, the model learns domain-specific features more effectively, producing more distinctive gradient patterns that GradAudit can exploit. In real-world scenarios, model owners often fine-tune on domain-specific copyrighted data to enhance specialized capabilities, making our method particularly valuable for real-world copyright verification.

\subsection*{Case Study: Detecting Copyrighted Material in VLLMs}
We conduct a case study examining potential copyright exposure in deployed VLLMs using real-world copyrighted content. We focus on Studio Ghibli, whose distinctive artistic style has become a focal point in recent AI copyright debates~\citep{ghibli2025lawsuit}. We collect 810 image-caption pairs from the Ghibli dataset on Hugging Face~\citep{ghibli2025dataset}, which contains frames from Studio Ghibli films paired with descriptive captions. This dataset represents a timely test case given ongoing legal and public discourse surrounding AI-generated content mimicking Ghibli's visual style.

We apply GradAudit and baseline methods to estimate the proportion of Ghibli data that may have been included in the training corpora of three VLLMs: MiniGPT-v2 (October 2023), Qwen2-VL (June 2024), and LLaVA-Med (August 2024). Although LLaVA-Med targets biomedical applications, it builds upon the general-domain LLaVA pre-trained on web-crawled data. Notably, Qwen2-VL and LLaVA-Med are more recent and more advanced than MiniGPT-v2~\citep{liu2024mmbench, duan2024vlmevalkit}. Since the probe dataset lacks confirmed negatives, we construct pseudo-negatives via image-text mismatching and apply a calibration procedure to obtain comparable estimates across all methods (details in Supplementary Information).

\begin{figure}[t!]
\centering
\includegraphics[width=\columnwidth]{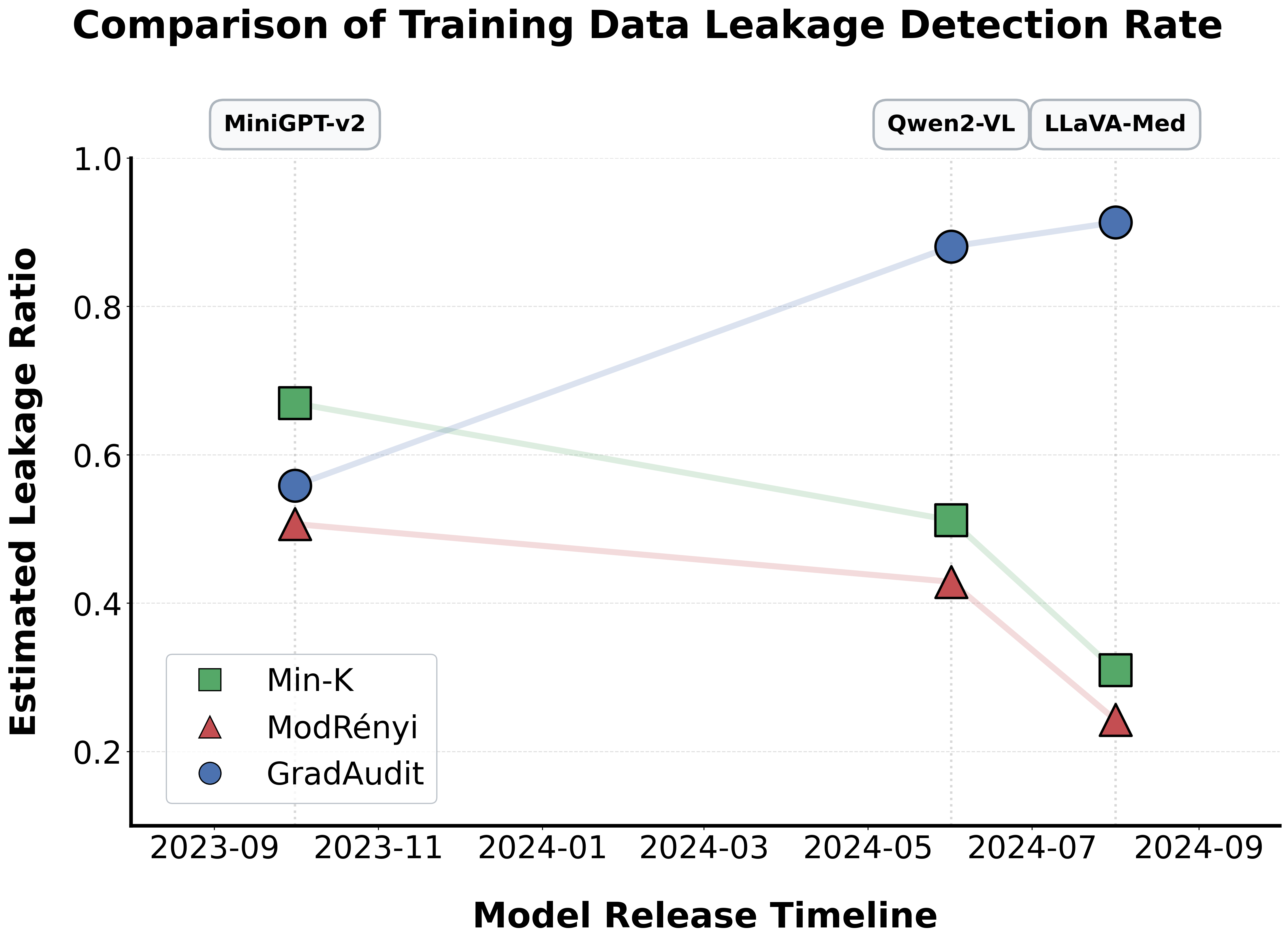}
\caption{\textbf{Estimating copyright exposure in VLLMs using Studio Ghibli data.} We apply GradAudit and baseline methods to 810 image-caption pairs from Studio Ghibli films to estimate potential training data inclusion across three VLLMs released between 2023 and 2024. GradAudit reveals an increasing trend of copyright exposure in more recent and more advanced models, while baseline methods (Min-K and ModRényi)  systematically underestimate the extent of copyrighted content in VLLM training data.}
\label{fig:leakage_detection}
\end{figure}

As shown in Figure~\ref{fig:leakage_detection}, GradAudit consistently estimates substantially higher leakage ratios than baseline methods across all three VLLMs, with the gap becoming more pronounced in more recent and more advanced models.  For Qwen2-VL, GradAudit estimates a leakage ratio of 0.881, compared to 0.512 for Min-K and 0.429 for ModRényi. For LLaVA-Med, the gap is even more pronounced: GradAudit estimates 0.913, while Min-K and ModRényi yield only 0.309 and 0.243 respectively. 

Moreover, GradAudit identifies a temporal pattern: estimated leakage ratios increase with model release date, rising from 0.558 for MiniGPT-v2 to 0.881 for Qwen2-VL and 0.913 for LLaVA-Med. One plausible explanation is that more recent VLLMs are trained on larger web-mined corpora~\citep{epoch2024datasetsizetrend}, which may increase exposure to copyrighted materials. In contrast, earlier models such as MiniGPT-v2 are likely to have a more limited exposure to the specific copyrighted materials.
Overall, the case study indicates that existing methods substantially underestimate unauthorized data usage in more recent, higher-capacity VLLMs, whereas GradAudit, which leverages internal parameter gradients, remains robust.

\begin{table*}[htbp]
\centering
\caption{\textbf{GradAudit performance under instruction variations on LLaVA-Med.} Five semantically equivalent prompts yield nearly identical results, demonstrating robustness to prompt formulation.}
\label{tab:instruction_robustness}
\begin{tabularx}{0.8\textwidth}{@{}Xcc@{}}
\toprule
\textbf{Instruction Prompt} & \textbf{AUC} & \textbf{TPR@5\%FPR} \\
\midrule
``Describe the medical image.'' & 0.6538 & 0.0987 \\
``What does this medical image show?'' & 0.6537 & 0.0962 \\
``Provide a detailed description of this medical scan.'' & 0.6535 & 0.0975 \\
``Analyze the medical imaging findings.'' & 0.6537 & 0.1000 \\
``What are the key observations in this medical image?'' & 0.6536 & 0.0987 \\
\midrule
\textbf{Mean $\pm$ Std} & \textbf{0.6537 $\pm$ 0.0001} & \textbf{0.0982 $\pm$ 0.0014} \\
\bottomrule
\end{tabularx}
\end{table*}

\subsection*{Robustness Analysis}
We evaluate GradAudit's robustness under two realistic scenarios: image perturbations and instruction variations. These experiments assess whether GradAudit maintains effective detection when the data used by auditors differs from the training data of the target model. Beyond these results, we also conduct sensitivity analyses examining the impact of key hyperparameters, including the noise masking threshold $\tau$, the number of gradient layers $K$, the choice of similarity function, and the reference data size. These analyses, provided in the Supplementary Information, demonstrate that GradAudit maintains robust performance across a range of parameter settings.

\subsubsection*{Robustness to Image Perturbations}
In practice, models are often trained on data collected from diverse web sources, where images may be compressed, resized, or otherwise degraded compared to the original copyrighted content. This quality mismatch poses a challenge for data auditing: even if copyright holders possess pristine original images, the target model may have been trained on corrupted versions obtained from unreliable sources. A robust auditing method must detect training data usage despite such discrepancies.

We simulate this scenario by training models on perturbed versions of images while auditing with original high-quality data. Specifically, we apply three common perturbation types to training images: JPEG compression, Gaussian blur, and Gaussian noise. We then evaluate whether GradAudit can still detect that these corrupted images were used during training, using only the original images for auditing. Experiments are conducted on both FashionGen and MedTrinity datasets using Qwen2-VL with LoRA fine-tuning.

\begin{figure}[t!]
\centering
\includegraphics[width=\columnwidth]{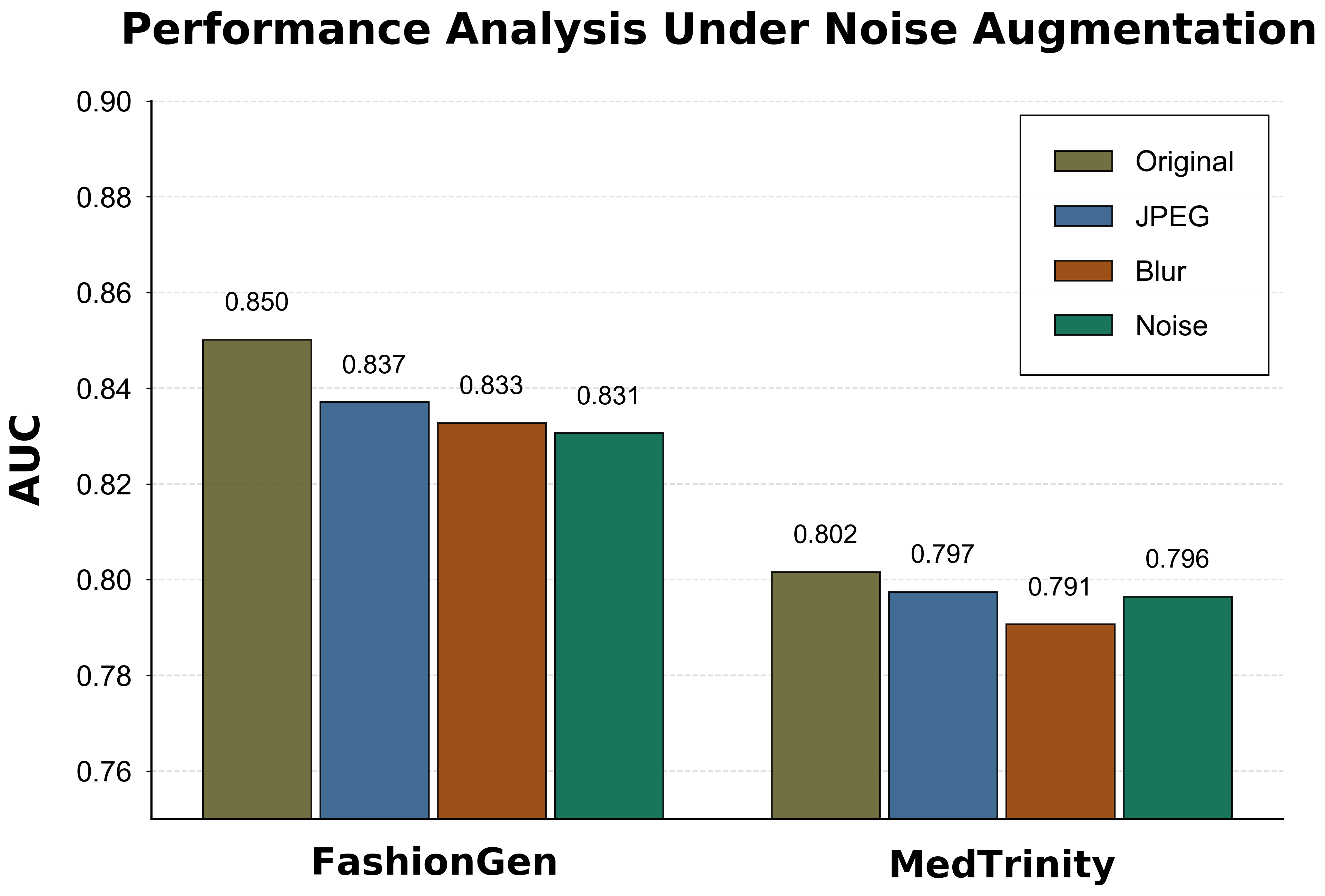}
\caption{\textbf{GradAudit robustness under image perturbations.} AUROC on FashionGen and MedTrinity under four conditions: original images (no perturbation), JPEG compression, Gaussian blur, and Gaussian noise. GradAudit maintains strong performance across all perturbation types, with AUROC consistently above 0.790 on both datasets.} 
\label{fig:robustness}
\end{figure}

As shown in Figure~\ref{fig:robustness}, GradAudit maintains strong performance across all perturbation types on both datasets. On FashionGen, AUROC decreases only marginally from 0.850 (original) to 0.837 (JPEG), 0.833 (blur), and 0.831 (noise)---a maximum drop of merely 1.9 percentage points. On MedTrinity, performance is similarly stable. These results demonstrate that GradAudit reliably detects training data usage even when there is substantial quality mismatch between training and auditing data.

This robustness can be attributed to the hierarchical nature of gradient signatures. When a VLLM learns from an image during training, it captures both low-level and high-level semantic information. The deep semantic representations learned by the VLLM are robust to pixel-level perturbations, as evidenced by the fact that compressed or degraded images can still be accurately recognized by the VLLM. The gradients that GradAudit extracts reflect this semantic memory, which remains largely preserved despite surface-level perturbations. Furthermore, our noise masking mechanism enables us to capture information that is more directly related to training, filtering out noise and focusing on discriminative signals. As long as the core semantic information is retained, gradient patterns remain discriminative for training data auditing.

\subsubsection*{Robustness to Instruction Variations}
Prompt engineering has become standard practice in VLLM deployment, where practitioners design task-specific instructions to optimize model performance. This raises a practical concern for data auditing: if GradAudit's effectiveness depends on using the exact prompts employed during training, its applicability would be severely limited, as such information is typically proprietary and unavailable to external auditors.

We evaluate GradAudit's robustness to instruction variations by testing five semantically equivalent but syntactically different prompts on LLaVA-Med (Table~\ref{tab:instruction_robustness}). GradAudit achieves remarkably consistent performance across all variants, with AUROC of $0.6537 \pm 0.0001$ and TPR@5\%FPR of $0.0982 \pm 0.0014$. This near-zero variance demonstrates that detection performance is essentially invariant to prompt formulation. The stability arises because gradient signals primarily reflect how well the model has memorized image-text associations---an intrinsic property of the training data itself rather than the specific query prompt.

This finding has important practical implications. First, auditors can employ standard, publicly available instruction templates without requiring knowledge of proprietary training prompts, significantly reducing the information requirements for effective copyright verification. Second, this robustness suggests that potential countermeasures involving prompt obfuscation (where model deployers deliberately use unusual instructions during training to evade detection) are unlikely to be effective, as the fundamental data memorization signatures persist regardless of prompt formulation.

\section*{Discussion}
\label{sec:discussion}
In this work, we present GradAudit, a gradient-based framework for detecting training data usage in Vision-Language Large Models. Our results demonstrate that gradient signatures provide a powerful discriminative signal for distinguishing training from non-training data, achieving up to 92.7\% AUROC on fine-tuning scenarios and 87.2\% on pre-training scenarios. These results substantially outperform existing methods that rely on model output features. Beyond overall performance gains, our experiments reveal several findings with implications for both the understanding of VLLM memorization and the practice of data auditing.

A central finding is that GradAudit captures genuine multimodal associations rather than individual modality membership. Our pairing analysis demonstrates that GradAudit achieves 85.5\% AUROC in distinguishing correctly paired training data from shuffled pairs where both images and texts appeared during training but not together. This capability is essential for multimodal copyright verification: it enables auditors to detect whether specific image-text pairs, the actual unit of copyrighted content, are used in training, rather than merely detecting whether individual images or texts appeared  in the training corpus. This is especially meaningful in practical settings such as healthcare, where medical images and corresponding diagnostic reports are inherently paired.

Our scaling analysis reveals that auditing performance improves consistently with both model size and fine-tuning data size, following patterns reminiscent of scaling laws observed in model capability research~\citealp{kaplan2020scaling}. This trend aligns with established understanding that larger models exhibit stronger memorization~\citealp{carlini2022quantifying}, producing more distinctive gradient signatures. Additionally, the particularly strong performance in fine-tuning scenarios (92.7\% AUROC) is especially significant because fine-tuning on domain-specific copyrighted data represents the most critical regime for both copyright infringement and privacy concerns~\citealp{yang2025adversarial}. Our case study involving Studio Ghibli data further shows that existing methods substantially underestimate the use of copyrighted material in more recent, higher-capacity VLLMs, while GradAudit remains effective for these newer models.

This work has several limitations that can be improved in the future. First, the method requires white-box access to model parameters and gradients, restricting applicability to scenarios with authorized access such as organizational audits, regulatory inspections, or court-ordered examinations. Although this requirement is fundamental to accessing internal optimization signals that provide substantial performance advantages, future work could explore privacy-preserving schemes such as secure multi-party computation or federated auditing approaches. Second, GradAudit requires reference datasets to identify training-sensitive gradient features and construct baseline representations. While this poses challenges when auditors lack confirmed training samples, publicly documented corpora widely adopted in VLLM training (e.g., LAION, Wikipedia, Common Crawl) can serve as reference datasets. Future work could develop zero-shot or few-shot variants requiring minimal reference data by exploiting architectural priors or meta-learning. Third, computing similarity across numerous gradient feature vectors incurs substantial computational overhead, potentially limiting applicability to extremely large-scale auditing scenarios. This could be addressed through adaptive feature selection methods, or approximate similarity metrics that more efficiently capture gradient patterns.

\section*{Methods}
\subsection*{Related Work}
\label{sec:related_work}
Data auditing (or training data detection) aims to determine whether specific samples were used to train a given model. For classification models, shadow model approaches~\citealp{shokri2017membership} and confidence-based metrics~\citealp{jia2019memguard} established foundational techniques. For language models, methods based on token-level predictive logits~\citealp{shi2024detecting}, divergence-based calibration~\citealp{zhang2024pretraining}, and contrastive decoding~\citealp{wang2025recall} have been proposed. White-box approaches that leverage gradients have shown strong performance~\citealp{nasr2018comprehensive}, demonstrating that gradient information provides discriminative signals for detecting training data. However, these methods target single-modality models and do not address the cross-modal challenges inherent to VLLMs. For vision-language models, existing methods such as ModRényi~\citealp{li2024membership}, M$^4$I~\citealp{hu2022m}, and embedding-based approaches~\citealp{wu2025image} rely on surface-level signals such as output entropy or similarity metrics, without examining internal optimization dynamics. As we demonstrate empirically, these methods achieve near-random performance on most VLLM configurations. Our work introduces the first gradient-based auditing framework specifically designed for VLLMs, leveraging distinctive gradient signatures of training data to achieve substantially improved detection accuracy.

\subsection*{GradAudit: Gradient-based Data Auditing Framework}
\label{sec:methodology}
\begin{figure*}[h]
    \centering
    \includegraphics[width=0.8\textwidth]{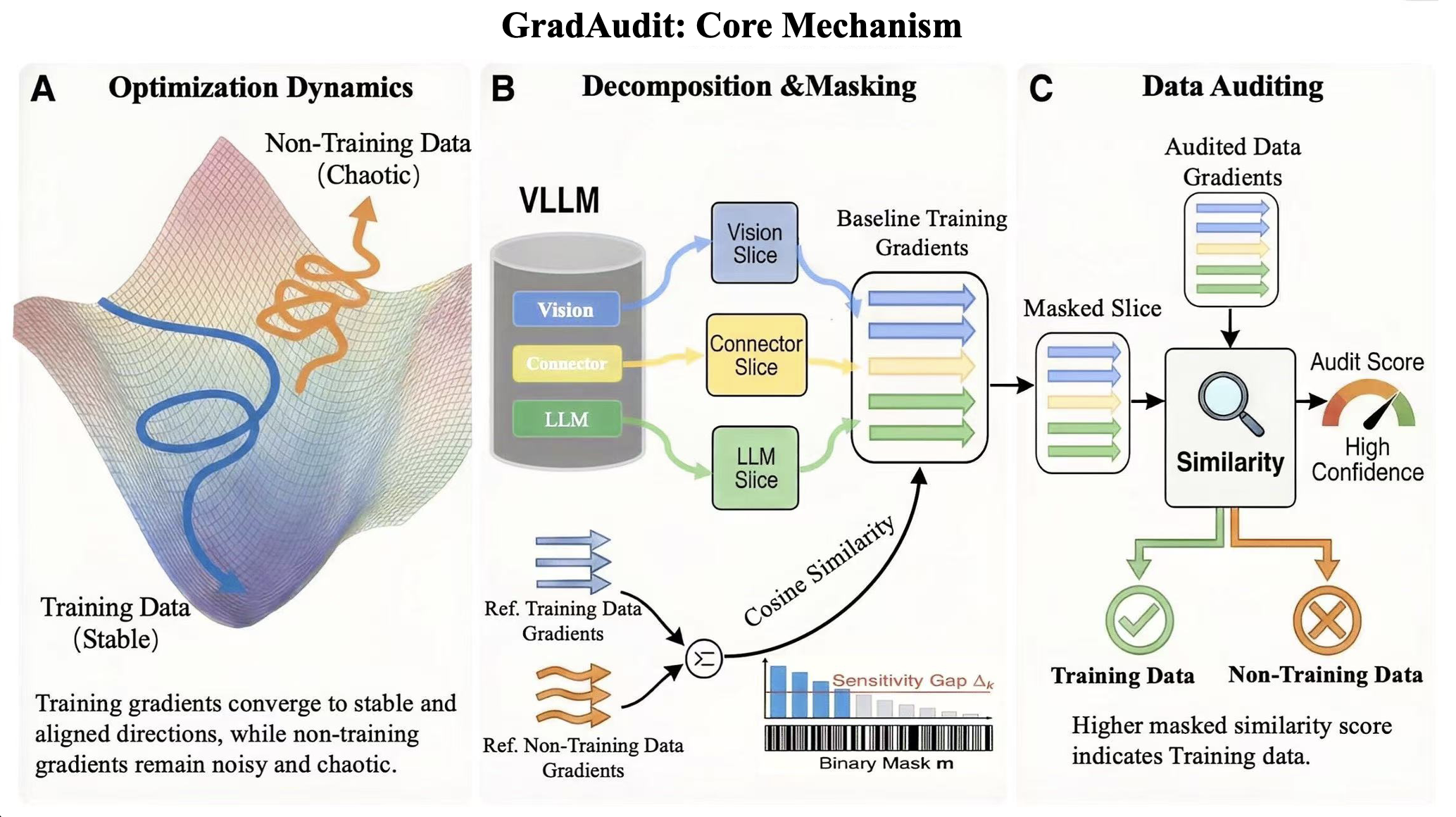}
    \caption{\textbf{Overview of the GradAudit framework.} Given an audited sample and reference data (comprising both training and non-training samples), GradAudit operates in three stages. (1) \textit{Gradient Feature Construction}: Parameter gradient matrices are decomposed into row and column slices, yielding functionally interpretable feature vectors. (2) \textit{Noise Feature Masking}: Reference data is used to compute sensitivity gaps across slices; slices with low discriminative power are masked out. (3) \textit{Data Auditing}: The masked similarity between the audited sample and reference training baseline determines the auditing decision.}
    \label{figures:gradaudit_framework}
\end{figure*}

GradAudit is a gradient-based data auditing framework that exploits the fundamental asymmetry in gradient behavior between training and non-training data. After optimization, gradients computed on training samples exhibit stable, well-aligned patterns, whereas gradients on non-training samples remain noisy and inconsistent. GradAudit leverages this asymmetry through three stages: (1) gradient feature construction via systematic slicing of parameter matrices, (2) noise masking to identify training-sensitive features, and (3) similarity-based auditing against reference training data.

\subsection*{Problem Definition} 
Given a Vision-Language Large Model $M$ and an audited sample $x = (x^{\text{img}}, x^{\text{txt}})$ comprising an image $x^{\text{img}}$ and associated text $x^{\text{txt}}$, the data auditing problem aims to determine whether $M$ was trained on $x$. This is formulated as a binary classification:
\begin{equation}
\label{eq:data_audit}
\mathcal{A}(M, x) \in \{0, 1\},
\end{equation}
where $\mathcal{A}(M, x) = 1$ indicates that model $M$ was trained on sample $x$.

\textbf{Threat Model.} We assume the auditor has white-box access to the target model, including its parameters and the ability to compute gradients for any input $x$. The auditor also has access to a small reference dataset comprising confirmed training and non-training samples. This assumption applies to: (i) internal audits within organizations verify compliance before model deployment~\citealp{wassie2024artificial}, and (ii) external audits where third parties are legally authorized to inspect model internals~\citealp{raji2022outsider}. The auditor does not require knowledge of training hyperparameters, data augmentation strategies, or the exact training procedure.

\subsection*{Gradient-based Feature Construction}
\label{sec:feature_construction}
A typical VLLM $M$ comprises a vision encoder, a language model, and cross-modal fusion modules, with combined parameters $\Theta$. For an input sample $x$, forward propagation produces a loss $\mathcal{L}(x)$, and backpropagation yields gradients $\nabla_{\Theta} \mathcal{L}(x)$ with respect to all parameters.

Our core hypothesis is that training data produce distinctive gradient signatures. During optimization, model parameters are iteratively updated along the gradient direction of training samples, causing parameters to converge to a region where gradients on training data become stable and flat~\citealp{feng2021inverse}. Non-training data, having never influenced the optimization trajectory, produce gradients with arbitrary directions in this converged region. This asymmetry renders gradients a discriminative signal for membership decision.

However, directly utilizing gradients across all parameters faces two critical challenges. First, modern large models contain billions of parameters, making exhaustive gradient analysis computationally prohibitive~\citealp{awais2025foundation}. Second, different parameters serve functionally distinct roles that cannot be captured through monolithic treatment~\citealp{ding2023parameter}. For instance, different rows of a vision encoder's parameter matrix $W \in \theta_v$ may encode diverse visual attributes such as spatial location, texture, and color~\citealp{bau2017network}. Therefore, directly using gradients across entire parameter matrices conflates signals from heterogeneous functions, obscuring which functional dimensions are most sensitive to training data.

To address these challenges, we propose a systematic slicing strategy that decomposes parameter gradient matrices into functionally specific units. This approach builds on prior work that utilizes gradient slicing to analyze dimension dependence to locate safety-critical~\citealp{xie2024gradsafe} and domain-sensitive~\citealp{tang2025gaprune} parameters in LLMs. Specifically, rather than treating each parameter matrix as an atomic block, we partition its gradient matrix along both row and column dimensions, where row slices correspond to output functional units and column slices correspond to input feature channels. This decomposition enables extraction of gradient features from functionally refined perspectives, facilitating identification of training-sensitive dimensions while dramatically reducing computational complexity.

Formally, for a parameter matrix $W \in \mathbb{R}^{N_{\text{out}} \times N_{\text{in}}}$ with gradient matrix $\nabla_{W} \mathcal{L}(x)$, we decompose it into row and column slices as follows:
\begin{equation}
\mathbf{g}_{\text{row}}^{(i)}(x, W) = \nabla_{W[i, :]} \mathcal{L}(x) \in \mathbb{R}^{N_{\text{in}}}, i = 1, \ldots, N_{\text{out}}, \label{eq:row_slice}
\end{equation}
\begin{equation}
\mathbf{g}_{\text{col}}^{(j)}(x, W) = \nabla_{W[:, j]} \mathcal{L}(x) \in \mathbb{R}^{N_{\text{out}}}, j = 1, \ldots, N_{\text{in}}, 
\label{eq:col_slice}
\end{equation}
where $\mathbf{g}_{\text{row}}^{(i)}(x, W)$ denotes the gradient feature vector corresponding to the $i$-th row of $W$, and $\mathbf{g}_{\text{col}}^{(j)}(x, W)$ denotes the vector corresponding to the $j$-th column of $W$. Aggregating across selected parameter matrices yields a set of $K$ gradient feature vectors:
\begin{equation}
\mathcal{G}(x) = \{\mathbf{g}_1(x), \mathbf{g}_2(x), \ldots, \mathbf{g}_K(x)\},
\label{eq:gradient_features}
\end{equation}
where each $\mathbf{g}_k(x)$ corresponds to a specific functional slice. In practice, we extract features from only the last few layers, as sensitivity analysis shows this achieves optimal performance while reducing computational cost (see Sensitivity Analysis).

\subsection*{Noise Feature Masking}
\label{sec:noise_masking}
Not all gradient features are equally informative for auditing. Generic components (e.g., early-layer feature encoders) exhibit low sensitivity to training data, while specialized components show high sensitivity~\citealp{luo2025empirical}. Additionally, parameter-efficient fine-tuning strategies~\citealp{hanparameter} modify only subsets of parameters, meaning specific samples may influence only particular slices. Since training details are typically proprietary, auditors cannot determine a priori which slices are training-sensitive. 

We address this challenge through a data-driven masking approach using reference data. Let $\mathcal{D}_{\text{ref}}^{\text{train}} = \{x_1^{\text{train}}, \ldots, x_N^{\text{train}}\}$ denote confirmed training samples and $\mathcal{D}_{\text{ref}}^{\text{non}} = \{x_1^{\text{non}}, \ldots, x_N^{\text{non}}\}$ denote confirmed non-training samples. We first compute the mean gradient vector of reference training data across all $K$ functional slices, establishing a baseline that characterizes the typical gradient signature of training influence:
\begin{equation}
\bar{\mathbf{g}}_{k}^{\text{train}} = \frac{1}{N} \sum_{i=1}^{N} \mathbf{g}_k(x_i^{\text{train}}), \quad k = 1, \ldots, K,
\label{eq:mean_gradient_mem}
\end{equation}
where $\mathbf{g}_k(x_i^{\text{train}})$ denotes the gradient feature vector of reference training sample $x_i^{\text{train}}$ on the $k$-th functional slice.

For each reference sample, we compute its cosine similarity to this baseline across all slices. Cosine similarity is chosen because it measures directional alignment, which  is more discriminative than magnitude-based metrics. For training samples:
\begin{equation}
\mathbf{s}_i^{\text{train}} = [s_{i,1}^{\text{train}}, \ldots, s_{i,K}^{\text{train}}] \in \mathbb{R}^K,
\label{eq:similarity_vector_mem}
\end{equation}
where
\begin{equation}
s_{i,k}^{\text{train}} = \frac{\mathbf{g}_k(x_i^{\text{train}}) \cdot \bar{\mathbf{g}}_{k}^{\text{train}}}{\|\mathbf{g}_k(x_i^{\text{train}})\| \cdot \|\bar{\mathbf{g}}_{k}^{\text{train}}\|} \in [-1, 1].
\label{eq:cosine_similarity_mem}
\end{equation}
An analogous computation is performed for each non-training sample $x_j^{\text{non}}$, yielding $\mathbf{s}_j^{\text{non}} \in \mathbb{R}^K$. We then aggregate similarity vectors across reference categories:
\begin{equation}
\bar{\mathbf{s}}^{\text{train}} = \frac{1}{N} \sum_{i=1}^{N} \mathbf{s}_i^{\text{train}} \in \mathbb{R}^K, \quad
\bar{\mathbf{s}}^{\text{non}} = \frac{1}{N} \sum_{j=1}^{N} \mathbf{s}_j^{\text{non}} \in \mathbb{R}^K.
\label{eq:mean_similarity_ref}
\end{equation}
To quantify the discriminative power of each functional slice $k$, we calculate the sensitivity gap, represented by the difference in mean similarity between reference training and non-training data.
\begin{equation}
\Delta_k = \bar{s}_k^{\text{train}} - \bar{s}_k^{\text{non}}, \quad k = 1, \ldots, K.
\label{eq:sensitivity_gap}
\end{equation}
Intuitively, a small $\Delta_k$ indicates negligible difference in baseline correlation between training and non-training data on slice $k$, suggesting the slice is either generic or unaffected by training and hence constitutes a noisy feature that should be masked. We construct a binary mask vector $\mathbf{m} = [m_1, \ldots, m_K] \in \{0, 1\}^K$, where each element is determined by comparing against a sensitivity threshold $\tau$:
\begin{equation}
m_k = \begin{cases} 1 & \text{if } \Delta_k > \tau \\ 0 & \text{if } \Delta_k \leq \tau \end{cases}.
\label{eq:mask_element}
\end{equation}
The resulting mask $\mathbf{m}$ filters out insensitive slices, retaining only features demonstrating genuine training sensitivity for subsequent auditing.

\subsection*{Data Auditing}
\label{sec:data_auditing}
After obtaining the mask vector $\mathbf{m}$ from the noise feature masking phase, we proceed to audit whether a given sample $x$ was used during training by computing the masked similarity between the audited sample and the reference training data baseline established in the previous section.

For the audited sample $x$, we first compute its cosine similarity to the baseline gradient vector $\bar{\mathbf{g}}^{\text{train}}$ across all functional slices:
\begin{equation}
    \mathbf{s}(x) = [s_1(x), \ldots, s_K(x)] \in \mathbb{R}^K,
    \label{eq:similarity_vector_audit}
\end{equation}
where each element $s_k(x)$ quantifies the similarity between the audited sample and the reference training baseline on the $k$-th functional slice:
\begin{equation}
    s_k(x) = \frac{\mathbf{g}_k(x) \cdot \bar{\mathbf{g}}_k^{\text{train}}}{\|\mathbf{g}_k(x)\| \cdot \|\bar{\mathbf{g}}_k^{\text{train}}\|} \in [-1, 1].
    \label{eq:cosine_similarity_audit}
\end{equation}

To focus on training-sensitive slices while suppressing noise, we apply the mask vector $\mathbf{m}$ element-wise to obtain the masked similarity vector:
\begin{equation}
    \mathbf{s}_{\text{masked}}(x) = \mathbf{s}(x) \odot \mathbf{m},
    \label{eq:masked_similarity}
\end{equation}
where $\odot$ denotes element-wise multiplication.

This operation effectively zeros out similarity values for noisy slices ($m_k = 0$) while preserving those for training-sensitive slices ($m_k = 1$).
We then compute the average masked similarity score by aggregating over all sensitive functional slices:
\begin{equation}
    \bar{s}_{\text{masked}}(x) = \frac{1}{\|\mathbf{m}\|_1} \sum_{k=1}^{K} s_k(x) \cdot m_k,
    \label{eq:avg_masked_similarity}
\end{equation}

Based on the average masked similarity score $\bar{s}_{\text{masked}}(x)$, we make the final auditing decision. 
The underlying intuition is straightforward: if the audited sample $x$ was used during training, its gradient features should align closely with the reference training baseline on sensitive slices, yielding a high similarity score. Conversely, if $x$ was not used during training, its gradient features should exhibit weaker alignment, resulting in a lower score. 
The decision is formalized as:
\begin{equation}
    \mathcal{A}(M, x) = \begin{cases}
        1 & \text{if } \bar{s}_{\text{masked}}(x) > \gamma \\
        0 & \text{if } \bar{s}_{\text{masked}}(x) \leq \gamma
    \end{cases},
    \label{eq:final_decision}
\end{equation}
where $\mathcal{A}(M, x) = 1$ indicates model $M$ is trained using data $x$ and $\gamma$ is a decision threshold. In practice, data auditing is typically evaluated using the Area Under the ROC Curve (AUROC), a threshold-independent metric that provides a comprehensive assessment of a method's discriminative performance.

\section*{Data availability}
All datasets used in this study are publicly available, and all experiments strictly adhere to their original licenses. LLaVA-Med instruction data and image URL metadata (referred to as PMC-OA in this paper) are obtained from the official repository \href{https://github.com/microsoft/LLaVA-Med}{https://github.com/microsoft/LLaVA-Med}. All other datasets, including ROCO, COCO, MedTrinity-25M, FashionGen, and Ghibli are accessed via Hugging Face Datasets \href{https://huggingface.co/datasets}{https://huggingface.co/datasets}.

\section*{Code availability}
All models used in this study are open-source. The source code for GradAudit and all experiments is publicly available at \href{https://github.com/tanghongyi0406/GradAudit}{https://github.com/tanghongyi0406/GradAudit}.

\bibliography{references}
\bibliographystyle{unsrtnat}

\section*{Author contributions}
Y.Y. led the research project. Y.Y. and Z.Z. conceived the idea of this work. H.T. implemented the models and conducted all experiments. Z.Z. and H.T. designed the experiments, analyzed the results, and led the writing of the manuscript. A.A. and Y.Y. contributed to the research design and manuscript writing. All authors reviewed and approved the final manuscript.

\section*{Competing interests}
The authors declare no competing interests.

\section*{Ethical and Societal Impact}
GradAudit is designed to strengthen copyright protection and data governance in the AI era by enabling verification of data usage compliance. 
We envision this capability serving multiple stakeholders: regulators seeking to enforce data protection requirements, content creators verifying whether their work was used without authorization, and AI developers demonstrating compliance with data usage agreements.

We acknowledge that gradient-based auditing techniques carry dual-use risks. While GradAudit requires white-box access (which provides an inherent safeguard by limiting use to authorized parties), related techniques could potentially be adapted to extract sensitive information from training data. To mitigate risks, our experiments exclusively use publicly available open-source models and anonymized datasets, introducing no new privacy vulnerabilities. We emphasize that our work aims to provide effective auditing tools for legitimate oversight of AI model data usage, ultimately contributing to a more accountable and trustworthy AI ecosystem. 

\section*{Additional information}
\label{sec:additional_information}
\begin{figure*}[htbp]
    \centering
    % --- Subfigure (a) Layer Depth ---
    \begin{subfigure}[b]{0.32\textwidth}
        \centering
        \includegraphics[width=\linewidth]{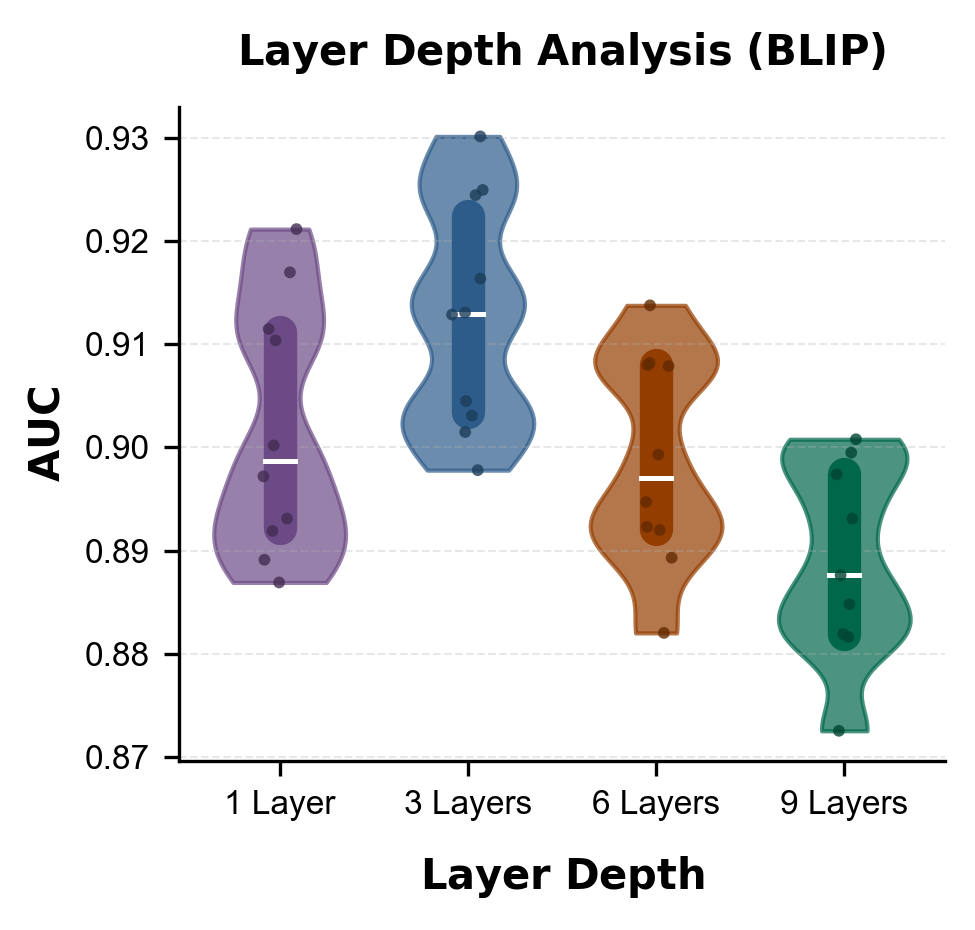}
        \caption{\textbf{Layer depth $K$.}}
        \label{figures:sensitivity_layer}
    \end{subfigure}\hfill
    % --- Subfigure (b) Threshold Tau ---
    \begin{subfigure}[b]{0.32\textwidth}
        \centering
        \includegraphics[width=\linewidth]{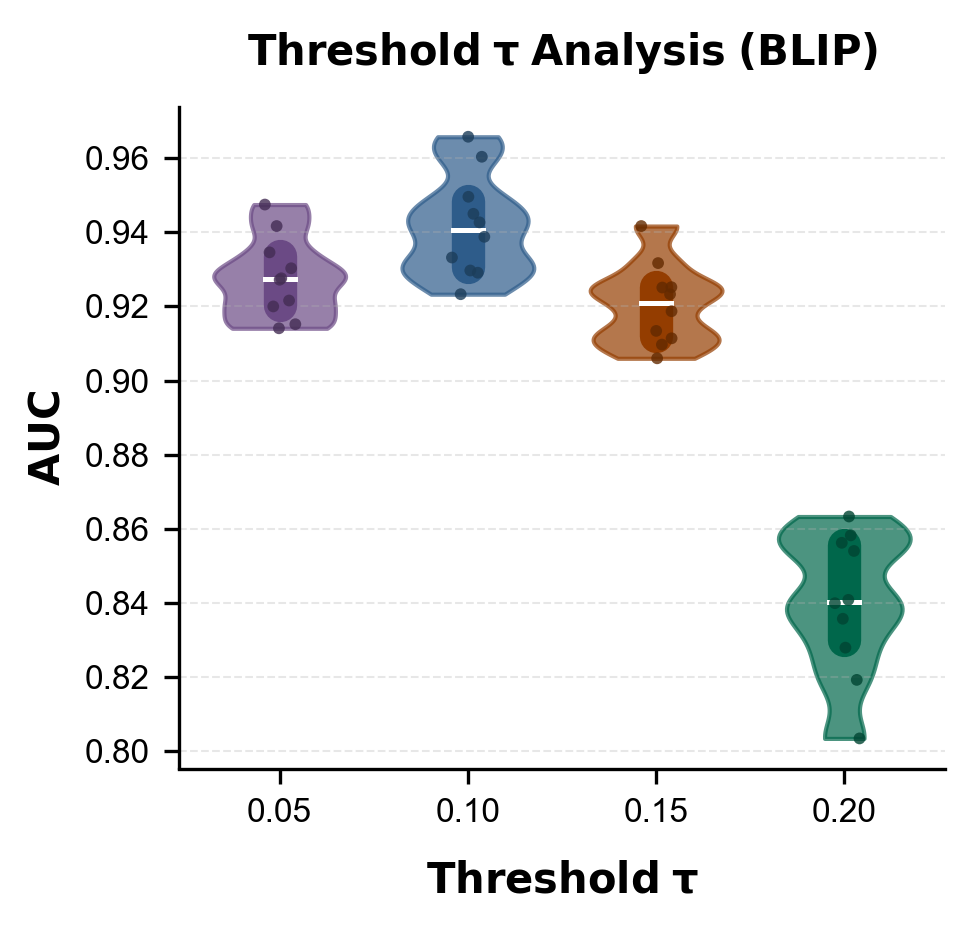}
        \caption{\textbf{Threshold $\tau$.}}
        \label{figures:sensitivity_tau}
    \end{subfigure}\hfill
    % --- Subfigure (c) Similarity Metric ---
    \begin{subfigure}[b]{0.32\textwidth}
        \centering
        \includegraphics[width=\linewidth]{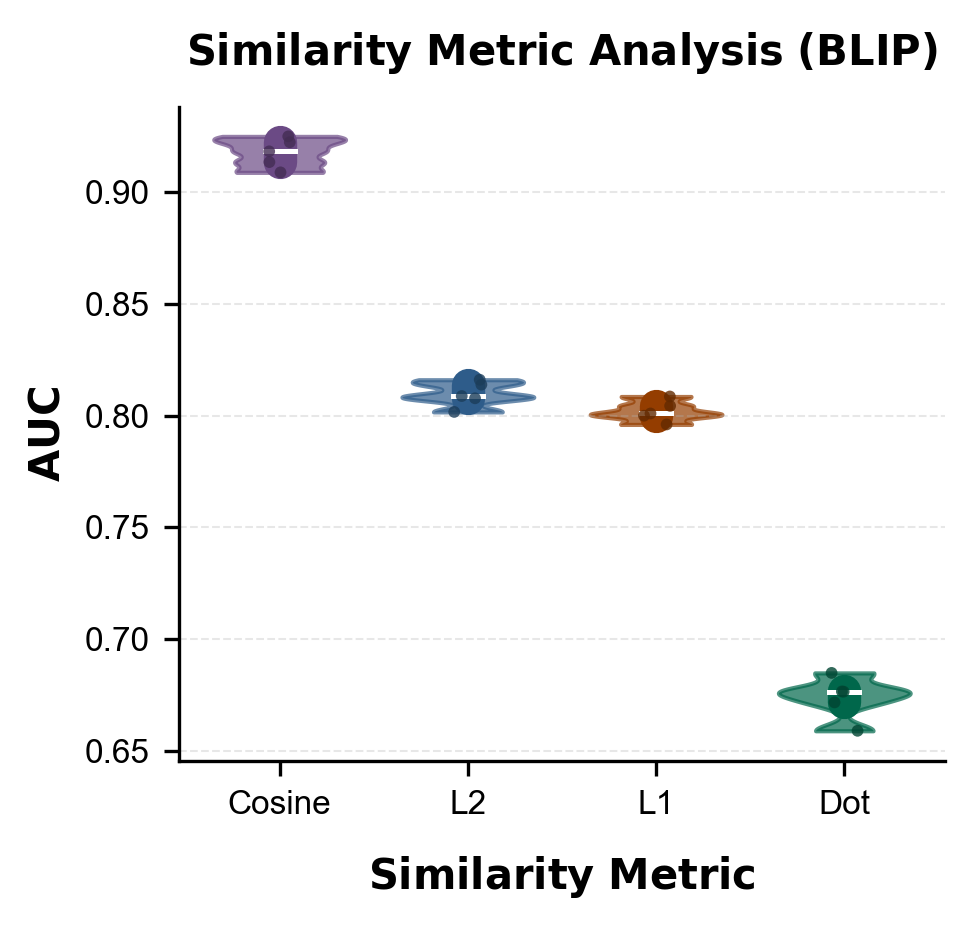}
        \caption{\textbf{Similarity metric.}}
        \label{figures:sensitivity_metric}
    \end{subfigure}
    
    \caption{\textbf{Sensitivity analysis of GradAudit hyperparameters on BLIP-ITM.} Violin plots illustrate the AUROC distributions across 10 independent runs for different hyperparameter configurations: 
    \textbf{(a) Layer depth $K$:} Performance peaks at $K=3$.
    \textbf{(b) Threshold $\tau$:} The auditing effectiveness is optimized at $\tau = 0.10$.
    \textbf{(c) Similarity metric:} Cosine similarity significantly outperforms distance-based metrics (L1, L2) and the raw dot product.}
    \label{figures:gradaudit_auc_sensitivity}
\end{figure*}

\begin{figure}[htbp]
\centering
\includegraphics[width=\columnwidth]{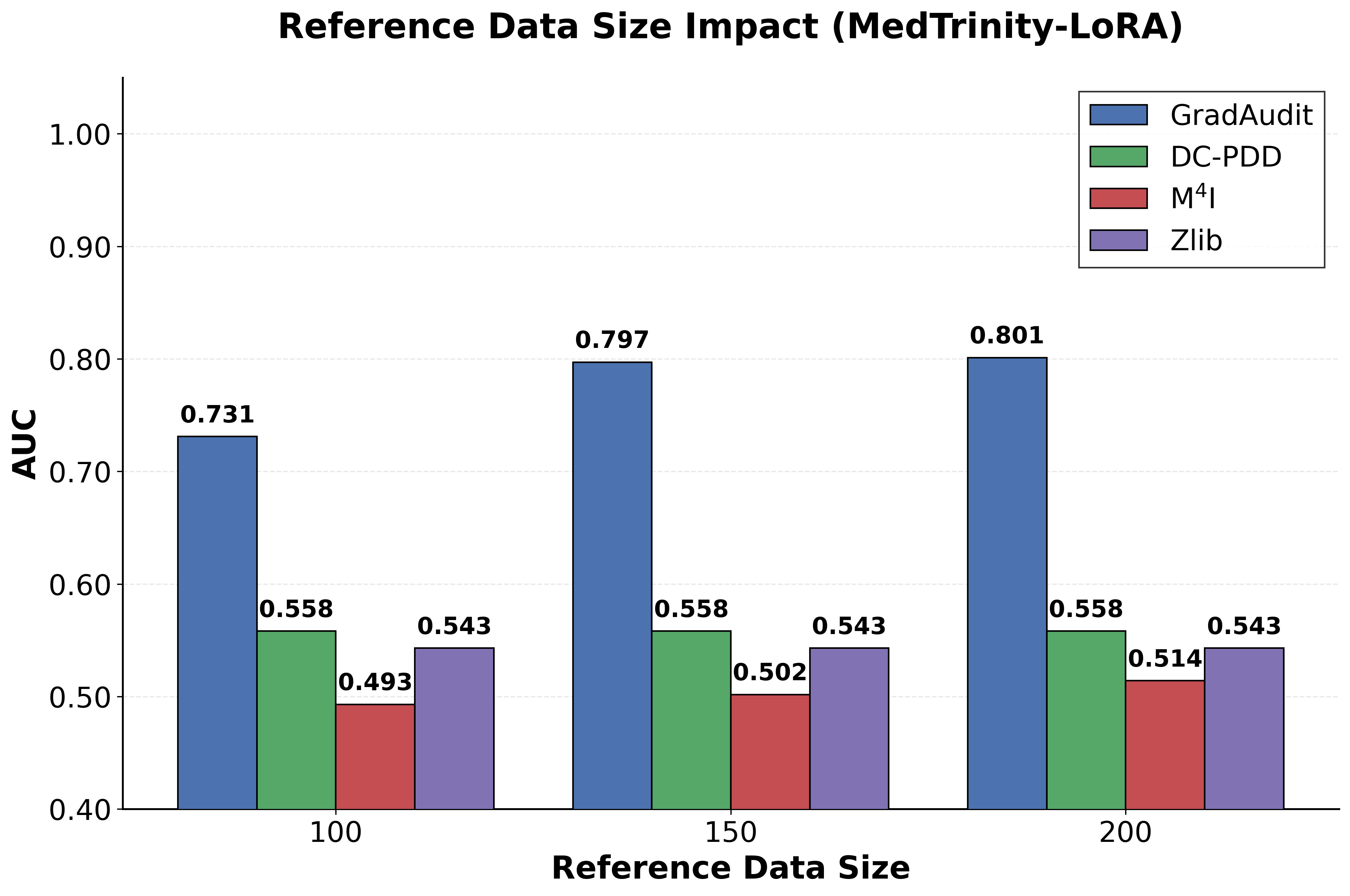}
\caption{\textbf{Impact of reference data size on auditing performance.} Even with only 100 reference samples, GradAudit substantially outperforms baselines ModRényi and GradNorm.}
\label{fig:reference_size}
\end{figure}

This supplementary material provides additional experimental details and analyses. We first present sensitivity analyses examining GradAudit's robustness to hyperparameter choices. We then describe the datasets, models, and fine-tuning configurations used in our experiments, followed by the complete GradAudit algorithm, detailed descriptions of baseline methods, and the Leakage Probability Calibration used in our case studies.

\subsection*{Appendix A: Sensitivity Analysis}
We investigate GradAudit's sensitivity to four key factors: the noise masking threshold $\tau$, the number of gradient layers $K$, the choice of similarity function, and the reference data size. Unless otherwise specified, experiments are conducted on BLIP-ITM with 200 training and 200 non-training samples from COCO. Results are shown in Figures~\ref{figures:gradaudit_auc_sensitivity} and~\ref{fig:reference_size}.

\textbf{Number of gradient layers $K$.}
We extract gradients from only the last $K$ layers, motivated by the observation that later layers encode more task-specific information and receive more direct gradient signals during backpropagation. As shown in Figure~\ref{figures:gradaudit_auc_sensitivity}(a), using 3 layers achieves the best performance (AUROC $0.914 \pm 0.007$), with marginal degradation for $K$=6 ($0.897 \pm 0.008$) and $K$=9 ($0.888 \pm 0.006$). This finding suggests that earlier layers contribute noise rather than discriminative signal, consistent with our noise masking strategy. An additional benefit of limiting $K$ is computational efficiency: reducing from $K$=9 to $K$=3 decreases gradient computation cost by approximately 3$\times$, making GradAudit more practical for real-world deployment. Based on these results, we adopt $K$=3 as the default value for all other experiments.

\textbf{Noise masking threshold $\tau$.}
The threshold $\tau$ controls which gradient features are retained based on their sensitivity gap between training and non-training reference data. As shown in Figure~\ref{figures:gradaudit_auc_sensitivity}(b), performance peaks at $\tau = 0.10$ (AUROC $0.943 \pm 0.013$) and declines sharply at higher thresholds (AUROC $0.840 \pm 0.018$ at $\tau = 0.20$). This sensitivity underscores the importance of noise masking: gradient features inherently contain substantial noise, and the threshold must balance retaining discriminative features while filtering uninformative ones. In practice, $\tau$ can be selected via cross-validation on a held-out subset of reference data. We find $\tau = 0.10$ provides robust performance and adopt it as the default across all experiments.

\textbf{Similarity function.}
The similarity function measures the similarity between gradient feature vectors extracted from the auditing data and reference data. We compare cosine similarity (our default), L2 distance, L1 distance, and dot product. As shown in Figure~\ref{figures:gradaudit_auc_sensitivity}(c), cosine similarity (AUROC $0.918 \pm 0.007$) substantially outperforms all alternatives: L2 ($0.810 \pm 0.006$), L1 ($0.802 \pm 0.005$), and dot product ($0.674 \pm 0.009$). The performance gap exceeds 10 percentage points, indicating that directional alignment of gradient vectors (rather than magnitude) is the critical signal for detecting training data. The poor performance of dot product suggests that gradient magnitudes vary substantially across samples and confound the training signal. 

\textbf{Reference data size.}
GradAudit requires reference data to compute gradient baselines and identify training-sensitive features. We investigate how the size of reference data affects auditing performance using Qwen2-VL-2B fine-tuned on MedTrinity. % with 3K training samples.
As shown in Figure~\ref{fig:reference_size}, GradAudit's performance improves with reference data size, achieving AUROC of 0.731, 0.797, and 0.801 for 100, 150, and 200 reference samples respectively. Performance plateaus beyond 150 samples, indicating diminishing returns from additional reference data. Importantly, even with only 100 reference samples, GradAudit (0.731) substantially outperforms both DC-PDD (0.558), M$^4$I (0.493) and Zlib (0.543), demonstrating robustness under limited reference data availability. Based on these results, we adopt 200 reference samples as the default configuration, which achieves near-optimal performance while remaining practically accessible. 

\subsection*{Appendix B: Datasets}

We evaluate GradAudit across five diverse datasets spanning medical and general domains, enabling comprehensive assessment of auditing performance under different data distributions and application scenarios. 

\noindent \textbf{PMC-OA.}  
A large-scale biomedical dataset containing image-text pairs from PubMed Central publications. It covers diverse medical imaging modalities (X-ray, CT, MRI, ultrasound, pathology) paired with clinical descriptions. PMC-OA serves as the primary training corpus for LLaVA-Med and BiomedCLIP.

\noindent \textbf{ROCO.}  
The Radiology Objects in Context dataset contains radiology images with captions from open-access biomedical literature. Sharing similar imaging characteristics with PMC-OA, ROCO is used as non-training data for evaluating auditing performance on biomedical models.

\noindent \textbf{COCO.}  
The Common Objects in Context dataset contains natural images with human-annotated captions depicting everyday scenes. Used as training data for BLIP-ITM evaluation.

\noindent \textbf{MedTrinity.}  
A Chinese multimodal medical dataset covering multiple imaging types (chest/abdomen X-ray, CT, ultrasound, endoscopy, fundus photography) with clinical reports in ShareGPT dialogue format. Used for fine-tuning experiments on Qwen2-VL.

\noindent \textbf{FashionGen.}  
A large-scale fashion dataset containing high-resolution product images paired with detailed textual descriptions including category, attributes, and style information. Used for fine-tuning experiments to evaluate auditing in general domains.

\begin{table*}[htbp]
\centering
\caption{\textbf{Fine-tuning validation on MedTrinity.} ROUGE-L scores confirm successful domain adaptation across training sizes and model scales, establishing that fine-tuned models are suitable targets for data auditing evaluation.}
\label{tab:medtrinity_lora}
\begin{tabular}{ccccc}
\hline
Data Size & Model Size & Base Score & After LoRA & Improvement \\
\hline
\multirow{2}{*}{1K} & 2B & 0.256 & 0.595 & +132.4\% \\
                    & 7B & 0.243 & 0.598 & +145.5\% \\
\hline
\multirow{2}{*}{3K} & 2B & 0.256 & 0.602 & +135.0\% \\
                    & 7B & 0.243 & 0.623 & +155.4\% \\
\hline
\multirow{2}{*}{5K} & 2B & 0.256 & 0.655 & +155.7\% \\
                    & 7B & 0.243 & 0.671 & +175.1\% \\
\hline
\end{tabular}
\end{table*}
\noindent \textbf{Training and Non-training Data Construction.}
We construct balanced auditing benchmarks with 1,000 training and 1,000 non-training samples per configuration. For pre-training scenarios, training samples are drawn from documented training corpora (PMC-OA for LLaVA-Med/BiomedCLIP, COCO for BLIP-ITM/MiniGPT-v2, ROCO for PubmedCLIP), while non-training samples come from held-out splits or distributionally similar datasets (e.g., ROCO for PMC-OA experiments). For fine-tuning scenarios (MedTrinity, FashionGen), both training and non-training samples are drawn from the same dataset, with strict exclusion of non-training samples from the fine-tuning process. This within-dataset sampling ensures that auditing performance reflects genuine membership detection rather than distributional artifacts.

\subsection*{Appendix C: Models}
We evaluate six VLLMs across three architectural paradigms: autoregressive generation, contrastive learning, and retrieval-based matching. These models demonstrate significant diversity in parameter scale (ranging from 151M to 8.2B), training objectives, and architectural designs, thereby enabling a comprehensive assessment of auditing effectiveness across heterogeneous model configurations.

\noindent \textbf{BiomedCLIP.}  
BiomedCLIP is a dual-encoder contrastive model for biomedical vision-language tasks. It employs a ViT-B/16 vision encoder and a PubMedBERT text encoder, totaling approximately 151M parameters. The model is trained on PMC-OA using symmetric InfoNCE loss, learning to align matched image-text pairs while separating non-matched pairs. Its contrastive objective provides a distinct evaluation setting compared to generative models.

\noindent \textbf{PubmedCLIP.}  
PubmedCLIP is a 151M-parameter dual-encoder (ViT-B/32 and distilled PubMedBERT) trained on ROCO to align radiology images and captions. Compared to BiomedCLIP, PubmedCLIP focuses specifically on radiology domain data, providing an additional evaluation setting for medical image-text auditing.

\noindent \textbf{BLIP-ITM.}  
BLIP-ITM is a dual-encoder retrieval model for image-text matching in general domains. The architecture consists of a ViT-B/16 vision encoder and a BERT-base text encoder. It is trained on COCO using a multi-task objective combining Image-Text Matching (ITM) classification and Image-Text Contrastive (ITC) learning, enabling both retrieval and semantic alignment.

\noindent \textbf{LLaVA-Med.}  
LLaVA-Med is an autoregressive generative VLLM designed for biomedical applications. The model comprises a CLIP ViT-L/14 vision encoder, a two-layer MLP projection module for cross-modal alignment, and a Vicuna-7B language decoder, totaling approximately 7.3B parameters. It is pre-trained on PMC-OA to generate detailed medical image descriptions. The autoregressive paradigm enables gradient computation through the complete generation process.

\noindent \textbf{MiniGPT-v2.}  
MiniGPT-v2 is a multi-stage generative VLLM with a modular architecture. It integrates an EVA-CLIP ViT-g/14 vision encoder, a BLIP-2 Q-Former for cross-modal feature extraction, and a LLaMA-7B language decoder, totaling approximately 8.2B parameters. Our evaluation focuses on the building MinGPT-v2 phase trained on the COCO dataset, which establishes foundational cross-modal mappings.

\noindent \textbf{Qwen2-VL.}  
Qwen2-VL is a modern generative VLLM featuring dynamic resolution processing and enhanced instruction-following capabilities. With approximately 2.3B parameters, it serves as the base model for our fine-tuning experiments using LoRA adaptation on MedTrinity and FashionGen. The smaller parameter count enables efficient experimentation while maintaining strong vision-language performance.

\subsection*{Fine-tuning Configuration}
To evaluate data auditing under fine-tuning scenarios (the predominant paradigm for deploying VLLMs in specialized domains), we conduct controlled experiments using Qwen2-VL-2B-Instruct as the base model. Fine-tuning is performed using LoRA (Low-Rank Adaptation), which introduces trainable low-rank matrices while keeping most pre-trained parameters frozen.

Our LoRA configuration uses rank $\text{r}=64$ with scaling factor $\alpha=128$, applied to all linear layers in attention and feed-forward modules. This modifies approximately 1\% of total parameters (~20M out of 2B), enabling efficient domain adaptation while preserving the base model's capabilities. Training uses the AdamW optimizer with learning rate $1\times10^{-4}$, weight decay 0.01, and batch size 16 (via gradient accumulation). Models are trained for 5 epochs with linear warmup (10\% of steps) followed by cosine decay, typically requiring 2-3 hours on a single A100 GPU for 1K-5K samples.

Table~\ref{tab:medtrinity_lora} validates that our LoRA fine-tuning procedure successfully adapts Qwen2-VL to the medical domain. All configurations show substantial improvements over base models (132--175\% relative gain), with performance scaling consistently with both training data size and model capacity. These results confirm that the fine-tuned models have genuinely learned from the training data, making them appropriate targets for evaluating whether GradAudit can detect such data usage. Similar validation was performed for FashionGen fine-tuning (results omitted for brevity), showing comparable improvement patterns.

\noindent \textbf{Computational Infrastructure.}
All experiments are conducted on NVIDIA A100 80GB GPUs. Memory optimization techniques include gradient checkpointing (30-40\% memory reduction) and Flash Attention (2-3× savings for self-attention). These enable training models up to 8B parameters on single GPUs while maintaining FP32 precision for accurate gradient computation.

\subsection*{Appendix D: GradAudit Algorithm}
\label{sec:algorithm}

\begin{algorithm}[t!]
\caption{\textbf{GradAudit Procedure}}
\label{alg:gradaudit}
\begin{algorithmic}[1]
\REQUIRE Target model $M$ with parameters $\Theta$, audited sample $x$, reference training data $\mathcal{D}_{\text{ref}}^{\text{train}} = \{x_1^{\text{train}}, \ldots, x_N^{\text{train}}\}$, reference non-training data $\mathcal{D}_{\text{ref}}^{\text{non}} = \{x_1^{\text{non}}, \ldots, x_N^{\text{non}}\}$, sensitivity threshold $\tau$, decision threshold $\gamma$
\STATE \textbf{\# Stage 1: Gradient feature extraction for reference data}
\FOR{each reference sample $x_i^{\text{train}} \in \mathcal{D}_{\text{ref}}^{\text{train}}$}
    \STATE Compute gradient features $\mathcal{G}(x_i^{\text{train}}) = [\mathbf{g}_1(x_i^{\text{train}}), \ldots, \mathbf{g}_K(x_i^{\text{train}})]$
\ENDFOR
\FOR{each reference sample $x_j^{\text{non}} \in \mathcal{D}_{\text{ref}}^{\text{non}}$}
    \STATE Compute gradient features $\mathcal{G}(x_j^{\text{non}}) = [\mathbf{g}_1(x_j^{\text{non}}), \ldots, \mathbf{g}_K(x_j^{\text{non}})]$
\ENDFOR
\STATE \textbf{\# Stage 2: Construct noise mask based on sensitivity gap}
\FOR{$k = 1$ to $K$}
    \STATE Compute training baseline: $\bar{\mathbf{g}}_{k}^{\text{train}} = \frac{1}{N} \sum_{i=1}^{N} \mathbf{g}_k(x_i^{\text{train}})$
    \STATE Compute mean similarity for training data: $\bar{s}_k^{\text{train}} = \frac{1}{N} \sum_{i=1}^{N} \cos(\mathbf{g}_k(x_i^{\text{train}}), \bar{\mathbf{g}}_{k}^{\text{train}})$
    \STATE Compute mean similarity for non-training data: $\bar{s}_k^{\text{non}} = \frac{1}{N} \sum_{j=1}^{N} \cos(\mathbf{g}_k(x_j^{\text{non}}), \bar{\mathbf{g}}_{k}^{\text{train}})$
    \STATE Compute sensitivity gap: $\Delta_k = \bar{s}_k^{\text{train}} - \bar{s}_k^{\text{non}}$
    \STATE Set mask: $m_k = \begin{cases} 1 & \text{if } \Delta_k > \tau \\ 0 & \text{otherwise} \end{cases}$
\ENDFOR
\STATE Construct mask vector: $\mathbf{m} = [m_1, \ldots, m_K]$
\STATE \textbf{\# Stage 3: Auditing based on masked similarity}
\STATE Compute gradient features for audited sample: $\mathcal{G}(x) = [\mathbf{g}_1(x), \ldots, \mathbf{g}_K(x)]$
\FOR{$k = 1$ to $K$}
    \STATE Compute similarity to baseline: $s_k(x) = \cos(\mathbf{g}_k(x), \bar{\mathbf{g}}_k^{\text{train}})$
\ENDFOR
\STATE Compute masked similarity score: $\bar{s}_{\text{masked}}(x) = \frac{1}{\|\mathbf{m}\|_1} \sum_{k=1}^{K} s_k(x) \cdot m_k$
\STATE Make auditing decision: $\mathcal{A}(M, x) = \begin{cases} 1 & \text{if } \bar{s}_{\text{masked}}(x) > \gamma \\ 0 & \text{otherwise} \end{cases}$
\RETURN $\mathcal{A}(M, x)$
\end{algorithmic}
\end{algorithm}

Algorithm~\ref{alg:gradaudit} describes the complete GradAudit procedure, which consists of three stages: gradient feature extraction, noise masking, and similarity-based auditing.

In the \textbf{gradient feature extraction stage} (Lines 1--7), we compute gradient representations for all reference samples. For each sample, we perform a forward pass through the model to compute the loss, then backpropagate to obtain gradients with respect to all parameters. These gradients are decomposed into $K$ functional slices using row and column partitioning of parameter matrices, yielding gradient feature vectors that capture different functional components of the model.

In the \textbf{noise masking stage} (Lines 8--15), we identify which gradient slices are informative for distinguishing training from non-training data. For each slice $k$, we first compute a baseline by averaging gradient features across all reference training samples. We then measure how well training and non-training samples align with this baseline using cosine similarity. The sensitivity gap, the difference in mean similarity between training and non-training data, quantifies each slice's discriminative power. Slices with gaps exceeding threshold $\tau$ are retained (mask value 1), while others are suppressed (mask value 0). This masking step is crucial for filtering out generic gradient components that do not reflect training influence.

\begin{table*}[t!]
\centering
\caption{\textbf{Comparison of baseline methods.} Methods are categorized by information source. ``Reference'' indicates whether the method requires reference data for calibration. ``VLLM-specific'' indicates whether the method was designed specifically for vision-language models.}
\label{tab:baseline_comparison}
\small
\begin{tabular}{llccc}
\toprule
\textbf{Category} & \textbf{Method} & \textbf{Info Source} & \textbf{Reference} & \textbf{VLLM-specific} \\
\midrule
\multirow{7}{*}{Output-based} 
& Loss Attack & Output loss & No & No \\
& Entropy & Output logits & No   & No \\
& Zlib    & Output loss   & Yes  & No \\
& Min-K\% & Output logits & No   & No \\
& Min-K++ & Output logits & No   & No \\
& DC-PDD  & Output logits & Yes  & No \\ 
& ModRényi& Output logits & No   & Yes \\
& Similarity & Output embeddings & No & Yes \\
& M$^4$I & Intermediate embeddings & Yes & Yes \\
\midrule
Gradient-based & GradNorm & Internal gradients & No & No \\
\midrule
\textbf{Ours} & \textbf{GradAudit} & \textbf{Internal gradients} & \textbf{Yes} & \textbf{Yes} \\
\bottomrule
\end{tabular}
\end{table*}

In the \textbf{similarity-based auditing stage} (Lines 16--23), we process the audited sample. We extract its gradient features using the same decomposition procedure, then compute cosine similarity between each slice and the corresponding training baseline. The final auditing score is the average similarity over retained slices only, weighted by the mask. A score exceeding threshold $\gamma$ indicates that the sample was likely used during training, as its gradient patterns closely align with those of confirmed training data.

\subsection*{Appendix E: Baseline Methods}

We compare GradAudit against nine representative baseline methods spanning two categories based on the information they utilize. Table~\ref{tab:baseline_comparison} summarizes the key attributes of all baseline methods.

\subsubsection*{Output-based Methods}

These methods analyze model outputs without accessing internal model states.

\noindent\textbf{Loss Attack}~\citealp{yeom2018privacy}: The most fundamental baseline based on the overfitting hypothesis. It assumes that the model's loss (e.g., cross entropy loss) on training data is significantly lower than on non-training data, as models tend to overfit to their training samples.

\noindent\textbf{Entropy}~\citealp{song2021systematic}: Measures prediction certainty by computing the entropy of the output probability distribution. Training data typically yields sharper distributions (lower entropy), reflecting the model's higher confidence on familiar samples.

\noindent\textbf{Zlib}~\citealp{carlini2021extracting}: Computes the ratio between an example's perplexity and its zlib compression entropy. This method normalizes the model's confidence by the inherent compressibility of the text, helping to distinguish between low perplexity due to training membership versus low perplexity due to simple, repetitive text patterns.

\noindent\textbf{Min-K\%}~\citealp{shi2024detecting}: Focuses on the most uncertain tokens rather than overall loss. It calculates the average log-likelihood of the $K\%$ tokens with the lowest probabilities, based on the observation that training data exhibits higher confidence even on difficult tokens.

\noindent\textbf{Min-K++}~\citealp{zhang2025min}: An improved version of Min-K\% that normalizes token probability with statistics of the categorical distribution over the entire vocabulary. Instead of using raw token probability, it computes a standardized score by comparing each token's log-probability against the mean and standard deviation of all vocabulary tokens at that position, providing a more informative membership signal.

\noindent\textbf{DC-PDD}~\citealp{zhang2024pretraining}: A divergence-based calibration method inspired by the divergence-from-randomness theory. It calibrates token probabilities by computing the cross-entropy between the token probability distribution predicted by the LLM and the token frequency distribution estimated from a reference corpus. 

\noindent\textbf{ModRényi}~\citealp{li2024membership}: Designed specifically for VLLMs, this method uses Rényi entropy of output logits as the membership signal. It computes the Rényi entropy at each token position and selects the largest K\% entropies for averaging, based on the assumption that training data yields lower entropy (higher confidence) in model outputs.

\noindent\textbf{Similarity}~\citealp{wu2025image}: A VLLM-specific method that measures feature consistency under input perturbations. It introduces noise (e.g., Gaussian blur) to inputs and computes cosine similarity between embeddings of original and perturbed outputs, assuming training data maintains more consistent representations under perturbations.

\noindent\textbf{M$^4$I}~\citealp{hu2022m}: A shadow model-based approach for multimodal models. It trains a binary classifier on feature difference vectors between image and text embeddings, exploiting the observation that training data exhibits tighter image-text alignment than non-training data.

\subsubsection*{Gradient-based Methods}
\noindent\textbf{GradNorm}~\citealp{nasr2018comprehensive}: Computes the $L_2$ norm of gradients with respect to model parameters. Based on the hypothesis that training data resides near local minima in the loss landscape, it expects smaller gradient norms for training samples compared to non-training samples.

\subsection*{Appendix F: Leakage Probability Calibration}

In the case study, the probe dataset contains only potential positive samples (copyrighted Ghibli content) without confirmed negatives. Directly applying a detection threshold may introduce bias, as different methods may have varying score distributions. To enable fair comparison across methods, we adopt a calibration approach based on differential statistics.

For each probe sample, we compute both the auditing score for the original image-text pair ($S_{\text{obs}}$) and a null score obtained by randomly mismatching the text ($S_{\text{null}}$). The difference $\Delta = S_{\text{obs}} - S_{\text{null}}$ captures the relative membership evidence attributable to the genuine pairing, thereby removing systematic biases stemming from dataset statistics or audit score distributions. To characterize the baseline distribution of $\Delta$ under the null hypothesis (no training exposure), we construct additional null pairs by computing $\Delta_{\text{null}} = S_{\text{null},1} - S_{\text{null},2}$ from two independent mismatchings, which reflects natural fluctuations due to gradient noise.

We then fit a two-component Gaussian mixture model to the observed $\Delta$ values:
$$p(\Delta) = (1-\pi) \cdot p_0(\Delta) + \pi \cdot p_1(\Delta)$$
where $p_0(\Delta) = \mathcal{N}(\mu_0, \sigma_0^2)$ represents the clean component (estimated from $\Delta_{\text{null}}$ and held fixed), $p_1(\Delta) = \mathcal{N}(\mu_1, \sigma_1^2)$ represents the leak component, and $\pi$ denotes the mixture weight. Parameters are estimated via Expectation-Maximization (EM) with the clean component fixed.

The calibrated leakage probability for each sample is computed as the posterior:
$$P(\text{leak}|\Delta) = \frac{\pi \cdot p_1(\Delta)}{(1-\pi) \cdot p_0(\Delta) + \pi \cdot p_1(\Delta)}$$

The dataset-level leakage ratio is obtained by averaging these posterior probabilities across all samples. This calibration procedure is applied consistently to GradAudit and all baseline methods.

\end{document}